\documentclass{article} %
\usepackage{colm2024_conference}

\usepackage{booktabs}
\usepackage{graphicx}
\usepackage{enumitem}
\usepackage{wrapfig}
\usepackage{algorithm}
\usepackage{algpseudocode}
\usepackage{multicol}

\usepackage{CJKutf8}

\usepackage{microtype}
\usepackage{amsmath}
\usepackage{amsfonts}
\usepackage{colortbl}
\usepackage[utf8]{inputenc}
\usepackage[T1]{fontenc}
\definecolor{lightgray}{rgb}{0.9,0.9,0.9}
\usepackage{caption}
\usepackage{subcaption}
\usepackage{xcolor}
\usepackage{setspace}
\usepackage{url}
\usepackage{multirow}
\usepackage{colortbl}
\usepackage{tabularx}
\usepackage{xltabular}
\usepackage{blindtext}
\usepackage{pgfplots}
\pgfplotsset{compat=1.18} 
\usepackage{tikz}
\usetikzlibrary{er,positioning,bayesnet}
\usepackage{makecell}
\usepackage{tipa}
\usepackage{siunitx}
\usepackage{nicefrac}
\usepackage{tocloft}
\usepackage{listings}
\usepackage[raster,skins]{tcolorbox} %
\usepackage{xltabular}
\usepackage{adjustbox}
\usepackage{xurl}
\usepackage{multirow}
\usepackage{threeparttable}
\usepackage{xcolor}

\usepackage{amsmath,amsfonts,bm}

\def\eqref#1{equation~\ref{#1}}

\def\1{\bm{1}}

\DeclareMathAlphabet{\mathsfit}{\encodingdefault}{\sfdefault}{m}{sl}
\SetMathAlphabet{\mathsfit}{bold}{\encodingdefault}{\sfdefault}{bx}{n}

\title{Qwen3.8-Omni: Towards Native Omni-Modal Agents }

\author{
\bf Qwen Team
}

\newcommand{\method}{Qwen3.8-Omni-Flash\xspace}

\newcommand{\plus}{Qwen3.5-Omni-Plus\xspace}
\newcommand{\flash}{Qwen3.8-Omni-Flash\xspace}
\newcommand{\realtime}{Qwen3.8-Omni-Flash-Realtime\xspace}

\begin{document}

\maketitle

\begin{abstract}

We introduce \method, a natively multimodal agentic model designed for real-world multimodal productivity. Compared with previous omni models, which primarily emphasized perception and interaction, \method substantially improves multimodal understanding and reasoning, as well as performance on long-horizon agentic tasks. These capabilities are supported by a native multimodal co-training strategy that preserves strong text-domain capabilities while facilitating the transfer of agentic capabilities from text to audio and video tasks. The model inherits the sparse mixture-of-experts (MoE) architecture of Qwen3.8-Next and extends the context window to one million tokens, supporting long-context multimodal reasoning and long-horizon planning. Together, these advances enable integration into production workflows as either a primary agent or a specialized sub-agent, supporting applications such as video editing, long-form audio and video translation, music-conditioned music video generation or movie generation, and video-based note or omni-skill creation. To address the lack of native audio and video support in existing agent harnesses, we release the open-source framework \textit{Qwen-MM-Plugins}\footnote{\url{https://github.com/QwenLM/Qwen-MM-Plugins}}, a lightweight plugin framework for multimodal productivity applications. We further frame real-time multimodal interaction as a system-level challenge requiring orchestration of context and memory management, tool use, and sub-agent delegation. Accordingly, we release \textit{Qwen-Live-Harness}\footnote{\url{https://github.com/QwenLM/Qwen-Live-Harness}}, an open-source framework for building responsive, real-time multimodal agents based on Qwen3.8-Omni. Extensive evaluations demonstrate that \method achieves strong performance across multimodal understanding, reasoning, long-horizon agentic execution, and video productivity tasks. These results, together with the accompanying open-source tools, support \method as a practical foundation for deploying natively multimodal agents in research and production settings.



\end{abstract}

\begin{figure}[tbh]
    \centering
    \makebox[0pt][c]{\includegraphics[width=1.0\textwidth]{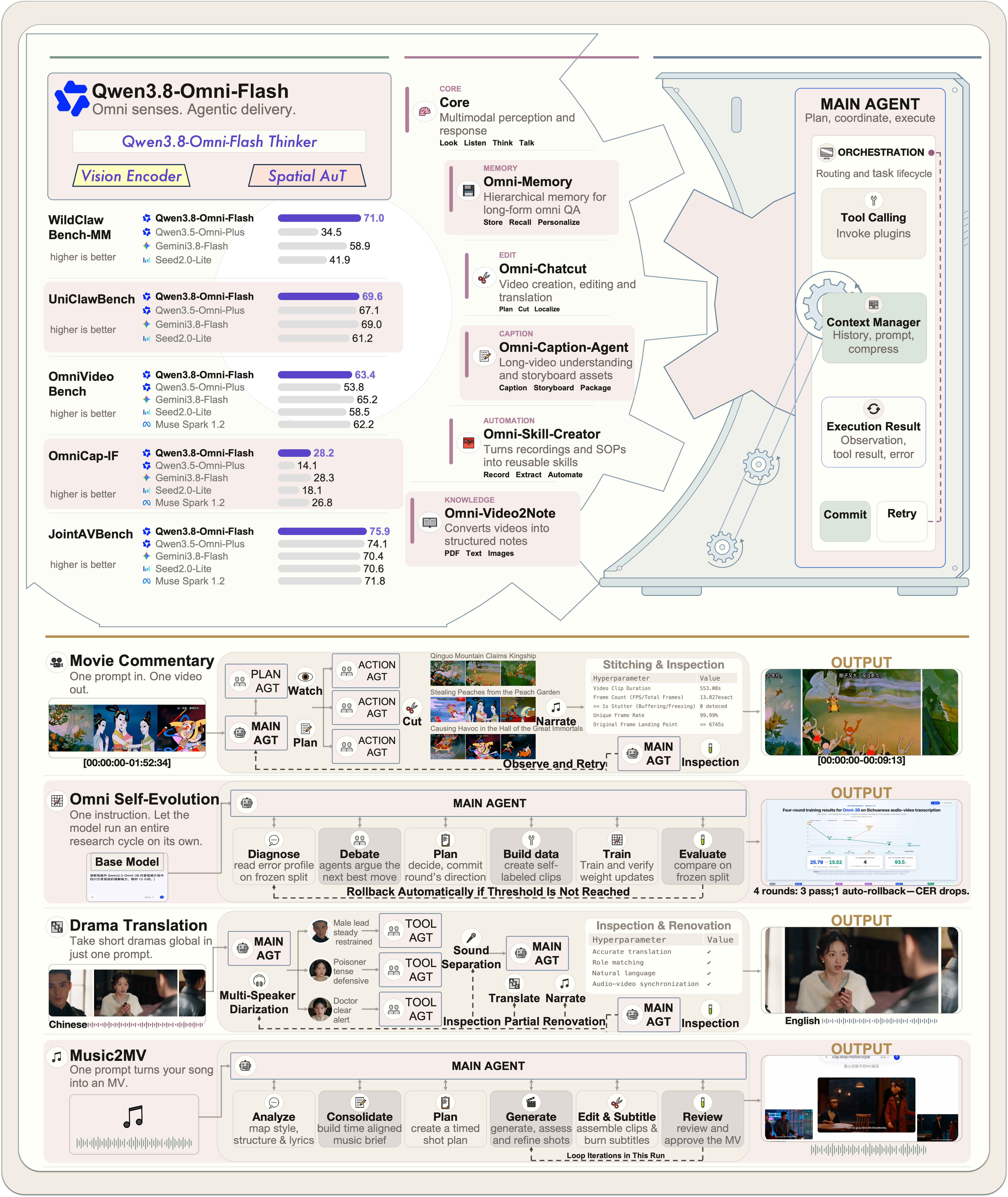}}
    \caption{\method~is a unified end-to-end model capable of processing multiple modalities, such as text, audio, image and video, and generating real-time text or speech response. Based on these features, \method~supports a wide range of tasks, including but not limited to voice dialogue, video dialogue, and audio-visual tool use.}
    \label{fig:intro}
\end{figure}


\section{Introduction}
\label{sec:intro}

Recent advances in reasoning models have made long-horizon agentic tasks increasingly feasible, with test-time computation emerging as a complementary axis of scaling~\citep{openai_o1_2024, anthropic_claude_thinking_2024, deepseekr1_2025, qwen38}. However, most existing agentic systems focus on software development, text-based knowledge work, graphical user interface interaction, or collaborative assistance, with multimodal capabilities used primarily for perception. This emphasis leaves a gap in support for video-centric, multimodal productivity workflows, such as agentic video editing, music video creation, and film production. These workflows require not only multimodal perception but also global planning, temporal reasoning, and the evaluation of coherence and quality in generated multimodal content. Moreover, long-form audiovisual understanding itself can benefit from agentic capabilities, including selective memory, narrative tracking, and targeted review of relevant segments. Motivated by these requirements and the limitations of existing systems, we investigate the potential of omni models to support agentic multimodal productivity.

To this end, we present \flash, Qwen’s latest native omnimodal MoE model, which supports reasoning over text, images, audio, and audiovisual content. The model follows the architecture of Qwen3.5-Omni Thinker~\citep{qwen35omni2026}, with a sparse MoE language backbone~\citep{qwen2026design} and an upgraded AuT~\citep{qwen35omni2026} designed to better capture spatial cues in audio. To develop the perceptual and reasoning capabilities required by multimodal productivity workflows, we first conduct native multimodal pretraining to strengthen unified perception and multimodal reasoning with long chains of thought. We then build on this foundation through post-training to enhance long-context reasoning and inference-time scaling. The post-training data and objectives emphasize long-horizon problems across text, vision, and audiovisual settings, spanning general reasoning, coding, and agentic workflows. To ground reasoning in environmental feedback, we complement static supervision with training in interactive environments, including executable code sandboxes and open harness-based interfaces. Finally, we apply multi-teacher distillation to consolidate domain-specialized and effort-specialized policies into a unified model for agentic multimodal productivity.

Our primary objective for this generation of omni models is to improve agentic productivity. Achieving this objective requires addressing three key challenges: (i) video inputs incur substantial storage and transmission costs and require large token budgets, making long-form processing expensive in agentic workflows; (ii) most existing harnesses lack support for streaming audio and video inputs within the main model context; and (iii) productivity-oriented applications of omni models remain underexplored.

To address the first challenge, we develop modules for information abstraction and on-demand access to multimodal content. Specifically, \textit{Omni-Caption} and \textit{Omni-Video2Note} convert videos into detailed captions and structured textual summaries, respectively; \textit{Omni-Memory} supports agentic lazy loading of audiovisual content; and \textit{Omni-Skill-Creator} transforms video tutorials or SOP-style content into executable skills that can be invoked by agent harnesses. Together, these modules reduce the need for dense processing of entire audiovisual inputs when abstracted information or selective retrieval suffices. To address the second challenge, we integrate these modules into \textit{Qwen-MM-Plugins}, which provides a lightweight interface through which existing agent harnesses can access audiovisual capabilities. To address the third challenge, the plugin includes ready-to-use productivity modules, such as \textit{Omni-Chatcut}, supporting applications including long-form audiovisual translation, film narration, and music video generation from audio tracks. All components are released as open source in \textit{Qwen-MM-Plugin}, and \method can operate either as a native multimodal main agent or as a sub-agent that invokes these tools.

Critically, \method supports sequences of up to one million tokens while maintaining text capabilities comparable to those of similarly sized text-only models and substantially improving omnimodal performance. Compared with its predecessor, Qwen3.5-Omni-Plus~\citep{qwen35omni2026}, it achieves an increase of more than 25\% in the average score across 29 evaluations spanning audio reasoning, audiovisual reasoning, and audiovisual agent tasks. Estimated API input costs per hour of audio and audiovisual content are reduced by more than 98\% and 93\%, respectively.

\section{Model Design}


\subsection{Architecture}
\method adopts the unified multimodal architecture, Thinker-Talker architecture, of the Qwen-Omni family~\citep{qwen2.5omni,qwen3omni,qwen35omni2026}. Compared with Qwen3.5-Omni-Plus~\citep{qwen35omni2026}, \method introduces several key improvements in scalability, alignment, and real-time interaction:
\begin{itemize}
\item The Thinker uses a hybrid sparse MoE language model to understand multimodal inputs and generate text for reasoning, dialogue, and tool use.
\item A vision encoder, Spatial AuT encoder, and AuT encoder provide visual, spatial-audio, and general-audio representations, respectively. Their outputs are projected into the Thinker's shared representation space.
\item Audio, spatial audio, and video representations include explicit timing information, helping the Thinker connect audio and visual events and reason about them across long input sequences.
\end{itemize}

\paragraph{Backbone} The Thinker builds on the hybrid sparse MoE language backbone of Qwen3.8-Next~\citep{qwen2026design}. Sparse expert activation increases model capacity while controlling the amount of feed-forward computation used by each token. For token mixing, the backbone combines Gated DeltaNet (GDN), which summarizes preceding context in a fixed-size recurrent state, with interleaved attention layers that retain direct access to context tokens.

The attention layers transition to Qwen Sparse Attention (QSA) through the warmup and sparse training stages. QSA uses a lightweight indexer to score compressed micro-block representations and select relevant context blocks, while core attention operates on the original tokens within the selected blocks. The resulting backbone combines recurrent processing with selective token-level retrieval over long multimodal sequences. Pretraining uses a native context window of 256K tokens throughout, followed by an extension to 1M tokens after post-training.

\paragraph{Multimodal Encoder} The Thinker uses three perception encoders for visual, spatial-audio, and general-audio inputs. The vision encoder, adopted from Qwen3.8-Next~\citep{qwen2026design}, processes images and sampled video frames.

\begin{figure}[tbh]
\centering
\includegraphics[width=0.65\textwidth]{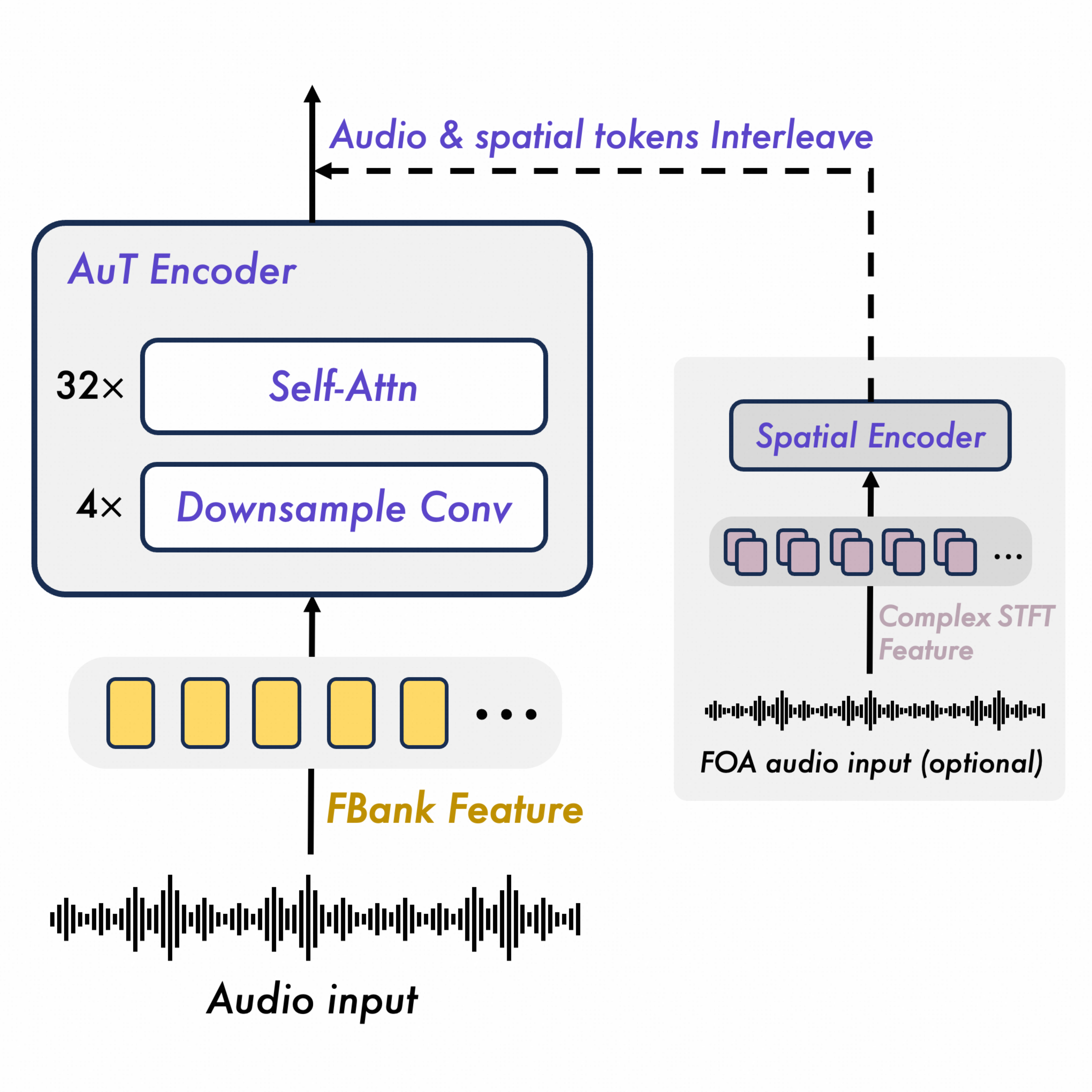}
\caption{Overview of the audio encoders in \method. The general AuT encoder extracts contextual audio representations at 6.25~Hz via convolutional downsampling and self-attention. A parallel Spatial AuT pathway processes multichannel spatial audio to capture directional and spatial cues.}
\label{fig:aut}
\end{figure}

For audio understanding, Qwen3.8-Omni-Flash combines a general-purpose AuT encoder for acoustic and linguistic content with a dedicated Spatial AuT encoder for spatial cues. The general-purpose AuT encoder~\citep{qwen35omni2026} uses a convolutional front end comprising four Conv2D blocks to downsample input acoustic features by a factor of 16, followed by temporal self-attention layers that produce contextualized audio representations. The encoder outputs tokens at 6,25~Hz, corresponding to approximately one token every 160~ms. AuT is pretrained on a sizable audio corpus, using transcriptions generated by a dedicated Qwen ASR model~\citep{qwen3asr} as supervision. The pretraining corpus covers more than 20 languages, with Chinese, English, and other languages represented in a ratio of 3.5:3.5:3. Dynamic attention-window training supports both streaming inference with cached context and offline audio understanding.

For spatial audio inputs, Spatial AuT provides a parallel processing pathway. A spatial interface first represents multichannel audio in first-order ambisonics (FOA) format. Spatial AuT then processes the complex-valued short-time Fourier transform (STFT) of the FOA channels, retaining both real and imaginary components to preserve interchannel amplitude and phase relationships. These relationships provide spatial cues associated with source direction, distance, and motion. A convolutional front end, temporal self-attention layers, and an output projection transform the spectral features into spatial audio representations. Modality-specific adapters project the outputs of both audio encoders and the vision encoder into the language backbone’s shared representation space, and all three encoders are aligned with the backbone during stage S1.

\subsection{Perception}
\label{sec:perception}

The Thinker converts text, images, audio, spatial audio, and video into a unified token sequence. Text is tokenized with the Qwen3.8-Next byte-level byte-pair encoding tokenizer, which uses a vocabulary of approximately 250K tokens~\citep{qwen2026design}. Images and dynamically sampled video frames are processed by the vision encoder. General audio, including audio extracted from video, is resampled to 16~kHz and converted into a 128-channel mel-spectrogram using a 25~ms window and a 10~ms hop before being encoded by AuT. For multichannel audio, Spatial AuT produces a parallel spatial representation that complements the general audio representation from AuT. The resulting visual, audio, and spatial-audio representations are projected into the Thinker's embedding space and arranged in temporal order for unified multimodal modeling. We retain explicit textual timestamps to expose the source timeline directly to the language backbone. The position IDs of spatial-audio embeddings are arranged analogously to those of visual patches at different spatial locations within a video frame: embeddings associated with the same temporal slice share a temporal index while retaining distinct spatial position indices. This layout preserves alignment with the audio stream while allowing the model to distinguish multiple spatial embeddings at each time step.

\section{Pretraining}

\method is pre-trained on a diverse dataset that encompasses multiple languages and dialects as shown in Table~\ref{table:languages} and modalities, including image--text, video--text, audio--text, video--audio, video--audio--text, and pure-text corpora. Following Qwen3.5-Omni-Plus~\citep{qwen35omni2026}, we employ a wider range of natural-language prompts to enhance both the generalization ability and instruction-following capabilities. To achieve robust performance across all modalities, our training strategy incorporates both unimodal and cross-modal data from the early pretraining stage.

\begin{table}[htbp]
\centering
\caption{\textbf{Supported languages and dialects in \method-Plus.}}
\vspace{-1mm}
\resizebox{\textwidth}{!}{
\begin{tabular}{lcp{11cm}}
\toprule
Modality & \# Varieties & Supported languages and dialects \\
\midrule
Text & 201 & See Qwen3.5 for the complete list of supported languages. \\
Speech Input & 113 & \textbf{74 languages:} Afrikaans, Arabic, Asturian, Azerbaijani, Basque, Belarusian, Bengali, Bosnian, Bulgarian, Cantonese, Catalan, Cebuano, Chinese, Croatian, Czech, Danish, Dutch, English, Esperanto, Estonian, Filipino, Finnish, French, Galician, Georgian, German, Greek, Hebrew, Hindi, Hungarian, Icelandic, Indonesian, Interlingua, Italian, Japanese, Javanese, Kannada, Kazakh, Korean, Kyrgyz, Lingala, Latvian, Lithuanian, Macedonian, Malay, Malayalam, Maltese, Maori, Marathi, Mongolian, Norwegian Bokmål, Norwegian Nynorsk, Oriya, Persian, Polish, Portuguese, Punjabi, Romanian, Russian, Serbian, Slovak, Slovenian, Spanish, Swahili, Swedish, Tajiki, Tamil, Telugu, Thai, Turkish, Ukrainian, Urdu, Uyghur, and Vietnamese.

\textbf{39 Chinese dialects:} Northeastern Mandarin, Guizhou dialect, Guangdong Cantonese, Henan dialect, Hong Kong Cantonese, Shanghainese, Shaanxi dialect, Tianjin dialect, Taiwanese Mandarin, Yunnan dialect, Anhui dialect, Fujian dialect, Gansu dialect, Guangdong Mandarin, Hubei dialect, Hunan dialect, Jiangxi dialect, Shandong dialect, Shanxi dialect, Sichuanese, Guangxi dialect, Hainan dialect, Chongqing dialect, Changsha dialect, Hangzhou dialect, Hefei dialect, Yinchuan dialect, Zhengzhou dialect, Shenyang dialect, Wenzhou dialect, Wuhan dialect, Kunming dialect, Taiyuan dialect, Nanchang dialect, Jinan dialect, Lanzhou dialect, Nanjing dialect, Hakka, and Southern Min. \\
Speech Output & 36 & \textbf{29 languages:} Chinese, English, German, Italian, Portuguese, Spanish, Japanese, Korean, French, Russian, Thai, Indonesian, Arabic, Vietnamese, Turkish, Finnish, Polish, Hindi, Dutch, Czech, Urdu, Tagalog, Swedish, Danish, Hebrew, Icelandic, Malay, Norwegian, and Persian.

\textbf{7 Chinese dialects:} Sichuanese, Beijing dialect, Tianjin dialect, Nanjing dialect, Shaanxi dialect, Cantonese, and Southern Min. \\
\bottomrule
\end{tabular}}
\label{table:languages}
\end{table}

We retain the temporal modeling strategy of Qwen3.5-Omni-Plus~\citep{qwen35omni2026}, including its audio-visual timestamp representation, without modification. In this generation, spatial audio is introduced as an additional input modality. Spatial AuT representations are temporally aligned with the existing audio and video representations, allowing spatial information to be incorporated into the unified multimodal sequence while preserving the established timestamp convention.

The pre-training of \method is structured into four distinct stages, all using a native sequence length of 262,144 tokens. In the first stage, we freeze the language model parameters and focus on aligning the vision encoder, AuT, and Spatial AuT using image--text, monaural audio--text, and multichannel audio--text data. In the second stage, we unfreeze all parameters and train with a wider range of multimodal data for more comprehensive learning. In the third stage, we keep the backbone fixed and warm up the QSA indexer using supervision from dense attention. In the final stage, we enable QSA and jointly optimize the backbone and indexer to adapt the model to sparse attention patterns:

\begin{enumerate}[label=(\arabic*)]
    \item \textbf{Encoder Alignment Stage (S1)}: During the initial pretraining phase, the LLM component of \method is initialized with parameters from Qwen3.8-Next~\citep{qwen2026design}, while the vision encoder is adopted from Qwen3.5 and the audio encoders are initialized with AuT and Spatial AuT. The three encoders are trained separately on the fixed LLM, initially focusing on their respective adapters before training the encoders.

    \item \textbf{General Stage (S2)}: The second phase of pretraining utilizes a large-scale dataset containing approximately 2.5 trillion tokens, with the following distribution across modalities: text (1.1 trillion), audio (0.7 trillion), image (0.35 trillion), video (0.15 trillion), and video--audio (0.3 trillion). The audio data include both monaural and multichannel audio. All model parameters are unfrozen during this stage. The introduction of more diverse multimodal data and tasks enhances the model's understanding and interaction capabilities in auditory, spatial-audio, visual, textual, and audio-visual information.

    \item \textbf{QSA Warmup Stage (S3)}: In the third phase, we keep the pretrained backbone fixed and train the QSA indexer using the full-attention distributions of the backbone as supervision. Dense attention remains active during this stage, allowing the indexer to learn block-level context selection before sparse attention is enabled.

    \item \textbf{QSA Stage (S4)}: In the final pretraining phase, we enable QSA in the corresponding attention layers and jointly train the backbone and indexer. This stage allows the model to adapt to sparse attention patterns while retaining its multimodal understanding and long-context capabilities.
\end{enumerate}
\section{Post-Training}\label{sec:post}

The post-training of the Thinker follows a two-stage pipeline comprising
\emph{Multi-Teacher Distillation} and \emph{Reinforcement Learning}.
The pipeline is designed to consolidate heterogeneous capabilities into a
unified model while mitigating cross-domain and cross-modal interference, and
to further improve response quality and interaction robustness through
reinforcement learning. The training data are serialized using the
ChatML~\citep{chatml} format and cover text-only, visual, audio, and
mixed-modality conversations.

\begin{itemize}

\item \textbf{Stage 1: Multi-Teacher Distillation.}
We first construct a collection of domain-specialized teacher models, each
initialized from the pre-trained Qwen3.8 base checkpoint and independently
optimized through supervised fine-tuning (SFT) and reinforcement learning
(RL). The specialists cover a broad range of capabilities, including general
instruction following, foundational reasoning, coding, agentic tasks, visual
understanding, and audio understanding. The specialist models are subsequently used to generate high-quality
domain-specific training trajectories. These trajectories are combined into a
unified multimodal training mixture and distilled into a single student model.
This formulation allows each specialist to provide targeted supervision in
its domain of expertise, while enabling the student to acquire these
capabilities within a shared parameterization. Compared with directly
fine-tuning the base model on a heterogeneous data mixture, specialist
distillation provides stronger and more consistent supervision, reduces
interference among modalities and task domains, and establishes a robust
initial policy for the subsequent reinforcement-learning stage.

\item \textbf{Stage 2: Reinforcement Learning.}
Starting from the distilled checkpoint, we perform unified reinforcement
learning across text, vision, audio, and mixed-modality tasks. Although
multi-teacher distillation equips the model with strong individual capabilities,
the resulting policy may still exhibit uneven response quality across input
modalities and suboptimal behavior in real-world interactions. In particular,
audio-conditioned queries remain more challenging than their text
counterparts, while long conversations may expose issues such as unintended
language switching, persona drift, and degradation in instruction following. To address these limitations, the RL task mixture jointly covers reasoning,
coding, agentic execution, multimodal understanding, audio-conditioned
dialogue, and multi-turn interaction scenarios. The reward signals evaluate
both task-level correctness and interaction quality. In addition to rewarding
accurate and helpful responses, the optimization encourages consistency
across input modalities, natural responses to spoken queries, stable language
and persona behavior, and reliable instruction following over extended
conversational contexts. For long-horizon agentic tasks, rewards are determined
by task outcomes and execution results rather than by the model's
self-reported completion. By jointly optimizing these objectives, the second stage transforms the
collection of distilled specialist capabilities into a coherent policy. The
resulting Thinker maintains strong performance across heterogeneous domains
while providing more consistent audio-conditioned responses and more stable
behavior in practical multi-turn interactions.

\end{itemize}
\section{Omni for Agentic Applications}

\subsection{Long Video Understanding}
Qwen3.8-Omni-Flash advances long-context audiovisual understanding across a range of capabilities, including active audiovisual perception, meeting understanding and follow-up task execution. These advances extend beyond longer temporal coverage to more precise evidence localization, deeper understanding, and more efficient execution.

\paragraph{Native Omni Agent for Long Video}
Conventional static approaches to long-video understanding feed all sampled frames and the accompanying audio in a single model invocation, even for videos lasting tens of minutes to several hours.
Increasing video duration leads to approximately linear growth in token consumption and greater difficulty in reasoning over the full input. 
However, reasoning may require evidence from only a few minutes of the video.
With native agentic capabilities, \method plans its analysis based on the query and invokes tools to acquire relevant audio and visual evidence.
Following a coarse-to-fine strategy, the model retrieves video and audio segments on demand through iterative tool calls.
This enables more efficient use of the limited token budget.
\method can also parallelize audio and visual analysis by delegating tasks to multiple subagents as needed.
This enables efficient analysis of videos spanning several hours within the limited context window of the main agent. 
By shifting long-video understanding from static processing to on-demand evidence gathering, native omni agent can improve accuracy while reducing token consumption.

\paragraph{Meeting Minutes for Multispeaker}
Multi-participant meetings are among the most complex audio-visual understanding scenarios: speakers take turns and overlap, while identities, references, and discussion topics continuously change. Traditional cascaded meeting analysis systems suffer from long and complex pipelines; even from the perspective of multi-speaker ASR alone, the sequential process of recognition, alignment, and diarization inevitably leads to severe error accumulation, further necessitating additional downstream models for content analysis. Qwen3.8-Omni-Flash jointly recognizes speakers across audio and video and natively supports up to one hour of audio-visual input. It can perform speaker segmentation, content transcription, and identity alignment end to end. Given a complete meeting video and a request, the model can map participant relationships, generate meeting minutes, identify action items, and analyze project risks, using visual information to resolve references and entity ambiguity in the audio. Combined with agents and tool use, it can also send emails, organize tasks, and even begin coding in response to meeting requirements, moving from understanding a meeting to acting on it.

\subsection{Content Creation}

Qwen3.8 Omni is taking audio-visual agents into a new stage: from understanding sounds and images to independently planning, calling tools, and delivering finished videos, bringing omnimodal intelligence into professional audio-visual content production workflows.

\paragraph{Music-to-MV}

We propose an agentic workflow for generating music videos from music tracks. Built on \flash and \textit{Qwen-MM-Plugins}, it offers a scalable solution for independent musicians and creators.
Creating visual content that complements a music track requires a comprehensive understanding of its lyrics, structure, and musical elements. With its strong music-understanding capabilities, \flash can analyze an input song and produce time-aligned lyric transcriptions, a global music caption, and localized descriptions of salient musical events organized along the song’s structural timeline. This musical evidence provides creative grounding for the agent to develop a narrative script and design individual shots. \flash can also review the generated shots for visual quality and correspondence with the creative plan. Together, \flash and \textit{Qwen-MM-Plugins} form an integrated workflow for end-to-end music-to-video production.

\paragraph{Drama Translation}
Traditional video translation typically involves multiple stages, including speech transcription, text translation, dubbing, audio mixing, and post-production editing. Coordinating these stages across different tools increases system complexity and makes it difficult to maintain consistency in character voices, dialogue duration, and visual pacing. Built on Qwen3.8-Omni-Flash and \textit{Qwen-MM-Plugins}, we develop an agentic workflow for short-drama translation, providing developers with a reference for designing automated video localization systems. Given a user request, the agent autonomously plans the localization process, selects and coordinates the required capabilities, and makes context-aware decisions throughout execution. It supports speaker-aware dialogue recognition, conversational translation, character-consistent voice cloning and dubbing, audio remixing, and quality assessment. By integrating these capabilities through autonomous planning and orchestration, the workflow enables the automated delivery of localized short dramas for international audiences.

\paragraph{Long-Form Movie Commentary}
For movies with runtimes of two to three hours, \method combines multimodal understanding with long-horizon planning to generate commentary videos autonomously end-to-end. The model analyzes narrative structure, key plot turning points, and character arcs across the full runtime, using this global context to guide commentary generation. It further orchestrates synthesized narration, original film dialogue, on-screen visuals, and background music into a coherent audiovisual composition, interleaving commentary with selected dialogue excerpts to preserve narrative coherence and emotional immersion. By adapting narration pacing and audio levels to the scene context, the model enables smooth transitions among commentary, original audio, and music, maintaining a balanced audio mix throughout the resulting video.

\subsection{Video for Research, Learning, and Automation}

Videos encode rich audiovisual information, yet relevant content is often dispersed across time, requiring repeated playback and manual navigation to locate and consolidate. Building on its omnimodal understanding capabilities, \method transforms video content into structured knowledge artifacts, including interactive research reports, instructional notes, and reusable agent skills. These artifacts consolidate essential information while preserving visual evidence and procedural context, making the underlying knowledge more accessible to both human users and AI agents. This enables video content to support research, learning, and automated task execution.

\paragraph{Omni-Deep Research}
\method supports video-centric omni-deep research, enabling in-depth investigation grounded in both video content and external knowledge. Given a video and a user query, the model analyzes the audiovisual content, identifies key research anchors such as claims, procedures, parameters, entities, and unresolved questions, and retrieves relevant information from web pages, documents, images, and other videos. It then integrates and cross-checks these multimodal sources with the original video, and organizes the results into an interactive HTML report that interleaves text with visual and video evidence and links research findings back to relevant moments in the source video. This capability is particularly useful for understanding complex concepts, verifying claims, reproducing procedures, diagnosing practical problems, and supporting informed decisions.

\paragraph{Video-to-Note}

Built on the powerful omni-modal understanding capabilities of \method, we developed Omni Video2Note, an agentic tool designed to help users extract information efficiently from large collections of tutorial videos, making it well suited for a wide range of instructional scenarios, including STEM education, everyday how-to content, and medical or caregiving training. By leveraging omni-modal inputs, it can reconstruct the instructional flow of a video and surface the most important points. For steps that benefit from visual context, it automatically selects the most informative frames from the source video. It can also proactively adjust the crop region to improve visual clarity and readability.

\paragraph{Skill Creator}
Video demonstrations convey both procedural knowledge and domain-specific expertise. To translate this knowledge into executable agent capabilities, we introduce Omni Skill Creator, an open-source component of \textit{Qwen-MM-Plugins} that constructs reusable skills from omnimodal content. The system extracts standard operating procedures (SOPs) from task demonstrations and distills tool usage, decision criteria, and key insights from expert instruction. The resulting skills undergo verification and evaluation, enabling knowledge captured in a single demonstration to be transformed into a reusable and shareable capability for automated task execution.

\subsection{Omni-Autoresearch}

With \method, we explore autoresearch: using a native omni model’s multimodal understanding to drive model development autonomously. We tasked it with improving Qwen2.5-Omni-3B~\citep{qwen2.5omni}’s Sichuan dialect speech recognition within 12 hours and delivering a usable model—a practical test of automating the expertise-intensive work needed to adapt smaller models for cost-sensitive applications. Qwen3.8-Omni-Flash independently selected the WenetSpeech-Chuan evaluation set, fixed the evaluation criteria, and established a baseline. It then listened directly to audio samples, diagnosed recognition errors, and constructed targeted training data. Across four rounds of experiments, it created 3,413 training examples and refined its approach using evaluation feedback, retaining effective changes and rolling back unsuccessful attempts. The smaller model’s character error rate fell from 25.79\% to 15.30\% on the same evaluation set, a relative reduction of approximately 40.7\%. This experiment demonstrates how native omni models can connect multimodal perception, experimental planning, and iterative training in an autonomous research workflow, helping smaller models acquire specialized capabilities for real-world applications.
\section{Evaluation}
\label{sec:experiment}

We organize the evaluation around the capabilities needed for native omni agents. Section~\ref{sec:multimodal-evaluation} establishes the model's general and modality-specific capabilities, spanning text, vision, audio, and audio-visual tasks. These include perception and reasoning as well as coding, office work, and visual agent tasks. Section~\ref{sec:omni-agent-evaluation} then examines omni agentic capabilities through multimodal task execution and agentic audio-visual evidence gathering. Together, the evaluations assess the model's capability profile, its progress over Qwen3.5-Omni-Plus~\citep{qwen35omni2026}, and the effect of agentic execution on understanding accuracy and token consumption.

\paragraph{Evaluation setup.}
For text and vision tasks, we compare against Qwen3.8-Flash~\citep{qwen3.8flash}, Qwen3.8-27B~\citep{qwen38}, Qwen3.7-Plus~\citep{qwen3.7plus}, and the additional baselines listed in the corresponding tables. For audio and audio-visual tasks, we include Qwen3.5-Omni-Plus~\citep{qwen35omni2026}, Gemini 3.8 Flash~\citep{gemini38flash}, Seed 2.0 Lite~\citep{seed2}\footnote{We use the API model \texttt{doubao-seed-2-0-lite-260428}.}, and Muse Spark 1.2~\citep{musespark12} where results are available. Benchmark-specific harnesses and scoring conventions are described in the table notes. Results involving tools reflect the model together with the stated execution setting. The capability groups below describe task domains rather than a distinction between tool-free and tool-assisted inference. We separately compare direct input interpretation with agentic evidence gathering in Section~\ref{sec:agentic-omni-understanding}. Bold values indicate the best result among the listed models for each metric, including ties; -- denotes an unavailable or inapplicable result.

\subsection{General and Multimodal Capabilities}
\label{sec:multimodal-evaluation}
We first examine four complementary capability domains. Text and visual evaluations cover both general reasoning and domain-specific agent tasks, while audio and audio-visual evaluations assess recognition, understanding, reasoning, and interaction through textual responses.

\subsubsection{Text Capabilities}
We evaluate software engineering, repository-level code generation, and office-oriented agent tasks using DeepSWE 1.1~\citep{deepswe1.1}, SWE-bench Pro~\citep{swebenchpro}, SWE-bench Multilingual~\citep{swebenchmultilingual}, NL2Repo-Bench~\citep{nl2repo-bench}, and CoWorkBench~\citep{qwen3.8flash}. General text capabilities are assessed using IFBench~\citep{ifbench}, GPQA Diamond~\citep{gpqa}, HLE~\citep{hle}, and LiveCodeBench v6~\citep{livecodebench}, covering instruction following, scientific and multidisciplinary reasoning, and competitive coding.


\begin{table}[H]
\centering
\caption{\textbf{Text capability evaluation of Qwen3.8-Omni-Flash and baseline models. The highest
scores are shown in bold.}}
\vspace{-.1in}
\label{tab:text_nonthink}
\begin{adjustbox}{max width=\textwidth}
\begin{threeparttable}
\small
\begin{tabular}{@{}lcccccc@{}}
\toprule
\multicolumn{1}{c}{%
\begin{tabular}[c]{@{}c@{}}
\textbf{Benchmark}
\end{tabular}}
& \begin{tabular}[c]{@{}c@{}}
\textbf{Qwen3.8-Omni-}\\
\textbf{Flash}
\end{tabular}
& \begin{tabular}[c]{@{}c@{}}
\textbf{Qwen3.8-}\\
\textbf{Flash}
\end{tabular}
& \begin{tabular}[c]{@{}c@{}}
\textbf{Qwen3.8-}\\
\textbf{27B}
\end{tabular}
& \begin{tabular}[c]{@{}c@{}}
\textbf{Qwen3.7-}\\
\textbf{Plus}
\end{tabular}
& \begin{tabular}[c]{@{}c@{}}
\textbf{DeepSeek-}\\
\textbf{V4-Flash-}\\
\textbf{0731}
\end{tabular}
& \begin{tabular}[c]{@{}c@{}}
\textbf{Claude-Opus-}\\
\textbf{4.6 (Max)}
\end{tabular}
\\ \midrule
\multicolumn{7}{c}{\textit{Coding and Agent}} \\ \midrule
DeepSWE 1.1           & 57.8          & \textbf{58.7} & 42.2 & 16.5 & 54.4          & --            \\
SWE-bench Pro          & \textbf{63.3} & 62.5          & 61.7 & 55.8 & 56.0          & 53.4          \\
SWE-bench Multilingual & 80.5          & \textbf{81.0} & 73.8 & 75.8 & --            & 77.5          \\
NL2Repo-Bench         & 48.9          & 48.1          & 42.3 & 41.1 & \textbf{54.2} & 47.6          \\
CoWorkBench           & \textbf{75.3} & 73.9          & 70.7 & 65.1 & 45.1          & 68.2          \\ \midrule
\multicolumn{7}{c}{\textit{General Text Capabilities}} \\ \midrule
IFBench                        & \textbf{81.5} & 81.3          & 79.5 & 79.1 & 79.2          & 62.5          \\
GPQA Diamond                   & 91.0          & \textbf{91.7} & 89.2 & 90.3 & 90.8          & 91.3          \\
HLE                   & 36.5          & 35.9          & 30.8 & 34.7 & 33.8          & \textbf{40.0} \\
LiveCodeBench v6               & \textbf{92.6} & 91.9          & 90.3 & 89.6 & 90.6          & 88.8          \\ \bottomrule
\end{tabular}
\begin{tablenotes}[flushleft]
\footnotesize
\item[1] DeepSWE 1.1: evaluated with the Claude Code and mini-SWE-agent harnesses,
temperature = 1.0, top\_p = 0.95, and a 256K context window.
We report the highest score across the two harnesses; notably,
Qwen3.8-Flash performs best on mini-SWE-agent.
\item[2] SWE-bench Pro: except for Claude-Opus-4.6 (Max), for which we report
the officially published score, all models are evaluated with the Claude Code
harness, temperature = 1.0, top\_p = 0.95, and a 256K context window.
Problematic tasks were corrected, and all models evaluated with our harness
were re-evaluated on the refined benchmark.
\item[3] SWE-bench Multilingual: evaluated with the mini-SWE-agent harness,
temperature = 1.0, top\_p = 0.95, and a 256K context window.
\item[4] NL2Repo-Bench: evaluated with the Claude Code harness.
To prevent reward hacking, we disable Bash commands that attempt to access
the specific repository, such as \texttt{pip download}, \texttt{pip install},
and \texttt{git clone}.
\item[5] CoWorkBench: an in-house cowork benchmark for evaluating long-horizon
office and productivity agent tasks across computer science, finance, law,
medical, and other productivity domains.
\item[6] HLE: judged by GPT-4o.
\end{tablenotes}
\end{threeparttable}
\end{adjustbox}
\end{table}

Table~\ref{tab:text_nonthink} shows that \method retains competitive text performance relative to Qwen3.8-Flash. It scores 63.3 on SWE-bench Pro, 48.9 on NL2Repo-Bench, and 75.3 on CoWorkBench, compared with 62.5, 48.1, and 73.9 for Qwen3.8-Flash. On DeepSWE 1.1 and SWE-bench Multilingual, it scores 57.8 and 80.5, respectively, slightly below 58.7 and 81.0. These results indicate that the model's multimodal capabilities coexist with strong coding and agent execution capabilities.

The same pattern extends to general text tasks: \method achieves 81.5 on IFBench and 92.6 on LiveCodeBench v6, the highest scores among the listed models, while obtaining 91.0 on GPQA Diamond and 36.5 on HLE. Performance varies across tasks rather than improving uniformly; for example, Claude-Opus-4.6 (Max) leads on HLE with 40.0.

\subsubsection{Visual Capabilities}
We assess visual agent capabilities with ClawEval-MM~\citep{claweval}, AndroidWorld~\citep{androidworld}, and Vision2Web~\citep{vision2web}. General visual evaluation covers embodied reasoning with ERQA~\citep{erqa}, long-video understanding with LVBench~\citep{lvbench}, real-world perception with RealWorldQA~\citep{realworldqa}, and mathematical and chart reasoning with MathVision~\citep{mathvision} and CharXiv (RQ)~\citep{wang2024charxiv}. For the latter two benchmarks, we report both the default setting and the additional CI setting.


\begin{table}[H]
\centering
\caption{\textbf{Visual capability evaluation of Qwen3.8-Omni-Flash and baseline models. The highest
scores are shown in bold.}}
\vspace{-.1in}
\label{tab:vision_text}
\begin{adjustbox}{max width=\textwidth}
\begin{threeparttable}
\small
\begin{tabular}{@{}lccccc@{}}
\toprule
\multicolumn{1}{c}{%
\begin{tabular}[c]{@{}c@{}}
\textbf{Benchmark}
\end{tabular}}
& \begin{tabular}[c]{@{}c@{}}
\textbf{Qwen3.8-Omni-}\\
\textbf{Flash}
\end{tabular}
& \begin{tabular}[c]{@{}c@{}}
\textbf{Qwen3.8-}\\
\textbf{Flash}
\end{tabular}
& \begin{tabular}[c]{@{}c@{}}
\textbf{Qwen3.8-}\\
\textbf{27B}
\end{tabular}
& \begin{tabular}[c]{@{}c@{}}
\textbf{Qwen3.7-}\\
\textbf{Plus}
\end{tabular}
& \begin{tabular}[c]{@{}c@{}}
\textbf{Claude-Opus-}\\
\textbf{4.6 (Max)}
\end{tabular}
\\ \midrule
\multicolumn{6}{c}{\textit{Agentic Vision Intelligence}} \\ \midrule
ClawEval-MM
& 60.4 $\mid$ \textbf{61.9}
& \textbf{64.4} $\mid$ 60.4
& 57.4 $\mid$ 56.9
& 57.4 $\mid$ 60.1
& 52.5 $\mid$ 54.7 \\
AndroidWorld                  & \textbf{87.1} & 84.5          & 81.9 & 81.0          & 62.0 \\
Vision2Web                    & 62.9          & \textbf{64.0} & 62.9 & 42.1          & --   \\ \midrule
\multicolumn{6}{c}{\textit{General Vision Capabilities}} \\ \midrule
ERQA                          & 71.0          & \textbf{72.3} & 65.5 & 69.8          & 40.8 \\
LVBench                       & \textbf{76.9} & 76.6          & 72.4 & 76.2          & 63.0 \\
RealWorldQA                   & 87.7          & \textbf{88.5} & 85.9 & 86.9          & 73.9 \\
MathVision                    & \textbf{91.8} & 90.6          & 90.0 & 90.3          & 65.5 \\
\multicolumn{1}{@{}r}{w/ CI}    & \textbf{96.2} & 95.7          & 94.6 & 88.4          & --   \\
CharXiv (RQ)                  & 83.5          & 84.6          & 83.7 & \textbf{85.8} & 66.0 \\
\multicolumn{1}{@{}r}{w/ CI}    & \textbf{91.4} & 90.6          & 90.2 & 85.9          & --   \\ \bottomrule
\end{tabular}
\begin{tablenotes}[flushleft]
\footnotesize
\item[1] ClawEval-MM results are reported as Pass@3 $\mid$ Average.
Pass@3 is the percentage of tasks passed in at least one of three trials;
Average is the mean score across the three trials.
\item[2] Vision2Web: scores are averaged over the frontend, webpage, and website
categories, using Claude Code and judged by gpt-5.4-2026-03-05.
\item[3] MathVision and CharXiv (RQ): the first row reports results without CI,
and the row labeled w/ CI reports results with CI.
A small number of incorrect MathVision ground-truth annotations were manually corrected.
Our model uses a fixed prompt requesting step-by-step reasoning and a final answer
in \texttt{\textbackslash boxed\{\}}. For other models, we report the higher score
from runs with and without this formatting instruction.
\end{tablenotes}
\end{threeparttable}
\end{adjustbox}
\end{table}

As shown in Table~\ref{tab:vision_text}, \method achieves 87.1 on AndroidWorld, exceeding Qwen3.8-Flash's 84.5. On ClawEval-MM, it obtains a higher average score (61.9 vs.\ 60.4), but a lower Pass@3 (60.4 vs.\ 64.4), illustrating the importance of distinguishing the two metrics. Its Vision2Web score of 62.9 is close to Qwen3.8-Flash's 64.0.

For general visual capabilities, \method achieves 76.9 on LVBench and 91.8 on MathVision in the default setting, the best results among the listed models. With CI, MathVision improves to 96.2 and CharXiv (RQ) improves from 83.5 to 91.4. Meanwhile, ERQA and RealWorldQA remain slightly below Qwen3.8-Flash. Overall, the model combines competitive visual perception with strong performance on selected visual agent and tool-assisted reasoning tasks.

\subsubsection{Audio Capabilities}
Our audio evaluation covers multi-speaker automatic speech recognition (ASR), general ASR, multilingual recognition, speech-to-text translation (S2TT), and instruction following, audio grounding and understanding, long-audio reasoning, music understanding, and audio-conditioned interaction. Multi-speaker ASR is assessed on AliMeeting Test~\citep{alimeet}, AISHELL-4~\citep{aishell4}, MagicData-RAMC~\citep{magicdata}, and MLC-SLM (en)~\citep{mlc-slm}, using DER and cpWER. WenetSpeech~\citep{DBLP:conf/icassp/ZhangLGSYXXBCZW22} complements these tasks with the Net and Meeting ASR subsets. The multilingual evaluation includes FLEURS-ASR and FLEURS-S2TT~\citep{Conneau2022FLEURSFL}, covering the 60 languages listed in Table~\ref{tab:audio_text}, together with OmniLingua-LongAudioASR, OmniLingua-AudioMaxIfe, and OmniLingua-MultiSpeaker from Omnilingua-Bench~\citep{omnilingua_bench_2026}, and MuLA-Bench~\citep{yang2026mula}. These additional benchmarks assess multilingual long-form transcription, spoken instruction following, multi-speaker transcription, and long-form audio understanding, respectively. We assess grounding with SpotSoundBench~\citep{sun2026spotsound}, general audio understanding with MMAU~\citep{sakshi2024mmaumassivemultitaskaudio}, MMAR~\citep{mmar}, and MMSU~\citep{mmsu}, and long-audio reasoning with LongAudioSpan~\citep{AudioSpan}, Vox-Infinity~\citep{cheng_2026_22868361} and LAMAR-Bench~\citep{wang_2026_22857938}. We further use MuchoMusic-RUL~\citep{zang2025you}, HumMusQA~\citep{HumMusQA}, and MusTBench~\citep{MusTBENCH} for music understanding, and Audio MultiChallenge~\citep{audio_multichallenge}, WildSpeech~\citep{zhang2025wildspeechbenchbenchmarkingendtoendspeechllms}, and VoiceBench~\citep{chen2024voicebench} for audio-conditioned interaction. All results in this subsection concern textual outputs from audio inputs.


\begin{table}[H]
\centering
\caption{\textbf{Audio capability evaluation of Qwen3.8-Omni-Flash and baseline models.
The best scores are shown in bold.}}
\vspace{-.1in}
\label{tab:audio_text}
\begin{adjustbox}{width=.9\textwidth}
\begin{threeparttable}
\small
\begin{tabular}{@{}lccccc@{}}
\toprule
\multicolumn{1}{c}{%
\begin{tabular}[c]{@{}c@{}}
\textbf{Benchmark}
\end{tabular}}
& \begin{tabular}[c]{@{}c@{}}
\textbf{Qwen3.8-Omni-}\\
\textbf{Flash}
\end{tabular}
& \begin{tabular}[c]{@{}c@{}}
\textbf{Qwen3.5-Omni-}\\
\textbf{Plus}
\end{tabular}
& \begin{tabular}[c]{@{}c@{}}
\textbf{Gemini 3.8}\\
\textbf{Flash}
\end{tabular}
& \begin{tabular}[c]{@{}c@{}}
\textbf{Seed 2.0 Lite}
\end{tabular}
& \begin{tabular}[c]{@{}c@{}}
\textbf{Muse Spark 1.2}
\end{tabular}
\\ \midrule
\multicolumn{6}{c}{\textit{Multi-Speaker ASR \& ASR}} \\ \midrule
AliMeeting Test
& \textbf{3.4} $\mid$ \textbf{17.2}
& 88.1 $\mid$ 89.6
& 72.6 $\mid$ 53.1
& 75.1 $\mid$ 76.1
& 93.7 $\mid$ 92.7 \\
AISHELL-4
& \textbf{2.8} $\mid$ \textbf{11.2}
& 100.0 $\mid$ 100.0
& 66.4 $\mid$ 56.9
& 64.8 $\mid$ 64.2
& 91.3 $\mid$ 86.0 \\
MagicData-RAMC
& \textbf{5.7} $\mid$ \textbf{14.1}
& 98.4 $\mid$ 97.1
& 67.9 $\mid$ 33.8
& 43.4 $\mid$ 35.1
& 82.1 $\mid$ 75.3 \\
MLC-SLM (en)
& \textbf{4.0} $\mid$ \textbf{14.2}
& 68.6 $\mid$ 63.9
& 60.8 $\mid$ 26.6
& 40.4 $\mid$ 45.5
& 74.3 $\mid$ 52.9 \\
WenetSpeech (Net)       & 4.8           & \textbf{3.7}  & 14.2          & 4.3  & 68.2 \\
WenetSpeech (Meeting)   & \textbf{4.6}  & 4.8           & 16.7          & 4.7  & 42.6 \\ \midrule
\multicolumn{6}{c}{\textit{Multilingual}} \\ \midrule
FLEURS-ASR     & 9.3           & \textbf{7.2}  & 7.9           & 32.1 & 23.6 \\
FLEURS-S2TT    & 31.8          & 32.2          & \textbf{33.0} & 24.8 & 28.8 \\
OmniLingua-LongAudioASR
& 95.56 $\mid$ 9.27
& 94.07 $\mid$ \textbf{6.89}
& 85.19 $\mid$ 12.87
& \textbf{97.78} $\mid$ 34.06
& \textbf{97.78} $\mid$ 84.41 \\
MuLA-Bench             & 72.60 & 61.00 & \textbf{73.43} & 58.06 & 34.04 \\
OmniLingua-AudioMaxIfe  & 86.50 & 83.10 & \textbf{87.90} & 60.70 & 82.20 \\
OmniLingua-MultiSpeaker
& \begin{tabular}[c]{@{}c@{}}\textbf{100.00} $\mid$ \textbf{43.25}\\\textbf{34.97} $\mid$ \textbf{20.15}\end{tabular}
& \begin{tabular}[c]{@{}c@{}}88.33 $\mid$ 84.46\\50.73 $\mid$ 47.61\end{tabular}
& \begin{tabular}[c]{@{}c@{}}97.00 $\mid$ 129.03\\44.58 $\mid$ 71.02\end{tabular}
& \begin{tabular}[c]{@{}c@{}}88.67 $\mid$ 99.28\\43.44 $\mid$ 50.42\end{tabular}
& \begin{tabular}[c]{@{}c@{}}98.67 $\mid$ 105.83\\77.19 $\mid$ 74.05\end{tabular} \\ \midrule
\multicolumn{6}{c}{\textit{Audio Grounding}} \\ \midrule
SpotSoundBench         & \textbf{67.2} & 64.2          & 39.7          & 59.6 & 16.9 \\ \midrule
\multicolumn{6}{c}{\textit{Audio Understanding}} \\ \midrule
MMAU                   & 81.8          & \textbf{81.9} & 76.9          & 77.2 & 63.5 \\
MMAR                   & \textbf{79.8} & \textbf{79.8} & 78.5          & 77.7 & 67.3 \\
MMSU                   & 82.1          & 83.0          & \textbf{83.3} & 80.2 & 59.9 \\ \midrule
\multicolumn{6}{c}{\textit{Long Audio Reasoning}} \\ \midrule
LongAudioSpan
& \textbf{82.7} $\mid$ \textbf{71.8} $\mid$ 48.2
& 74.4 $\mid$ 49.8 $\mid$ 45.1
& 79.3 $\mid$ 65.5 $\mid$ \textbf{64.6}
& --
& -- \\ 
Vox-Infinity
& \textbf{74.7} $\mid$ 68.3 $\mid$ 60.8
& 56.8 $\mid$ 64.6 $\mid$ 58.1
& 69.2 $\mid$ 58.4 $\mid$ \textbf{64.2}
& 69.9 $\mid$ \textbf{69.3} $\mid$ 59.1
& 69.3 $\mid$ 63.9 $\mid$ 48.1 \\ 
LAMAR-Bench
& \textbf{77.8} $\mid$ \textbf{62.2} $\mid$ \textbf{85.3}
& 64.1 $\mid$ 21.6 $\mid$ 54.3
& 71.1 $\mid$ 45.5 $\mid$ 71.2
& 43.4 $\mid$ 9.3 $\mid$ 58.2
& 34.1 $\mid$ 12.1 $\mid$ 33.4 \\ \midrule
\multicolumn{6}{c}{\textit{Music Understanding}} \\ \midrule
MuchoMusic-RUL         & \textbf{72.6} & 71.6          & 53.7          & 61.7 & 40.1 \\
HumMusQA               & \textbf{75.8} & 75.5          & 71.2          & 66.0 & 63.3 \\
MusTBench              & \textbf{50.6} & 49.1          & 40.3          & 44.0 & 29.4 \\ \midrule
\multicolumn{6}{c}{\textit{Audio Interaction}} \\ \midrule
Audio MultiChallenge   & 71.5          & 57.6          & \textbf{71.9} & 63.4 & 57.9 \\
WildSpeech             & 74.3          & 75.7          & \textbf{76.4} & 74.5 & 73.4 \\
VoiceBench             & 91.6          & \textbf{92.9} & 92.3          & 84.1 & 79.8 \\ \bottomrule
\end{tabular}
\begin{tablenotes}[flushleft]
\footnotesize
\item[1] AliMeeting Test, AISHELL-4, MagicData-RAMC, and MLC-SLM (en)
results are reported as DER $\mid$ cpWER (lower is better for both metrics).
WenetSpeech and FLEURS-ASR results are reported as WER (lower is better);
FLEURS-S2TT results are reported as BLEU (higher is better).
For audio understanding, music, and interaction scores, higher is better.
\item[2] FLEURS: ASR and S2TT evaluation results both cover the following
60 languages: Chinese (Mandarin), English, Cantonese, Arabic, German,
French, Spanish, Portuguese, Indonesian, Italian, Korean, Russian, Thai,
Vietnamese, Japanese, Turkish, Hindi, Malay, Dutch, Urdu, Norwegian,
Swedish, Danish, Hebrew, Finnish, Polish, Icelandic, Czech, Filipino,
Persian, Greek, Afrikaans, Asturian, Belarusian, Bulgarian, Bengali,
Bosnian, Catalan, Cebuano, Estonian, Galician, Gujarati, Croatian,
Hungarian, Javanese, Kazakh, Kannada, Kyrgyz, Latvian, Macedonian,
Malayalam, Marathi, Punjabi, Romanian, Slovak, Slovenian, Swahili,
Tajik, Azerbaijani, and Ukrainian.
\item[3] OmniLingua-LongAudioASR results are reported as success rate ($\uparrow$)
$\mid$ WER ($\downarrow$), both in percent.
\item[4] MuLA-Bench and OmniLingua-AudioMaxIfe: higher scores are better.
\item[5] OmniLingua-MultiSpeaker: each cell reports success rate ($\uparrow$)
$\mid$ tcpWER ($\downarrow$) on the first line and cpWER ($\downarrow$)
$\mid$ DER ($\downarrow$) on the second line. All four metrics are in percent;
the best value for each metric among the listed models is shown in bold.
\item[6] LongAudioSpan results are reported as
Accuracy $\mid$ Rubric $\mid$ Chain;
the highest score for each metric is shown in bold.
\item[7] Vox-Infinity results are reported as
Ultra Multi-Turn $\mid$ Personal Monologues $\mid$ Beyond-Semanti;
the highest score for each metric is shown in bold.
\item[8] LAMAR-bench results are reported as
Track A $\mid$ Track B $\mid$ Track C;
the highest score for each metric is shown in bold.
\end{tablenotes}
\end{threeparttable}
\end{adjustbox}
\end{table}

\paragraph{Multi-speaker recognition and audio understanding.}
The largest improvements over Qwen3.5-Omni-Plus occur in multi-speaker ASR. On AliMeeting Test, DER and cpWER decrease from 88.1 and 89.6 to 3.4 and 17.2, respectively. \method also achieves the lowest values for both metrics on AISHELL-4, MagicData-RAMC, and MLC-SLM (en). In audio grounding, SpotSoundBench improves from 64.2 to 67.2. On LongAudioSpan, Accuracy increases from 74.4 to 82.7 and Rubric from 49.8 to 71.8, although the Chain score of 48.2 remains below Gemini 3.8 Flash's 64.6. The model also leads the listed baselines on MuchoMusic-RUL, HumMusQA, and MusTBench, with scores of 72.6, 75.8, and 50.6, respectively.

\paragraph{Multilingual capabilities.}
The expanded multilingual evaluation reveals improvements in long-form understanding, instruction following, and multi-speaker transcription over Qwen3.5-Omni-Plus. On MuLA-Bench, \method improves from 61.00 to 72.60, close to Gemini 3.8 Flash's 73.43. OmniLingua-AudioMaxIfe increases from 83.10 to 86.50, compared with 87.90 for Gemini 3.8 Flash. On OmniLingua-MultiSpeaker, \method achieves a 100.00\% success rate and the lowest tcpWER, cpWER, and DER among the listed models, at 43.25\%, 34.97\%, and 20.15\%, respectively. Recognition and translation remain more mixed. On OmniLingua-LongAudioASR, success rate improves from 94.07\% to 95.56\%, while WER increases from 6.89\% to 9.27\%. FLEURS-ASR likewise yields a higher error rate of 9.3 than Qwen3.5-Omni-Plus (7.2) and Gemini 3.8 Flash (7.9), and FLEURS-S2TT scores 31.8 compared with 32.2 and 33.0. These results distinguish gains in multilingual understanding and multi-speaker processing from the remaining gaps in general transcription and translation.

\paragraph{General ASR and interaction.}
\method obtains the lowest WenetSpeech (Meeting) error rate of 4.6, while its WenetSpeech (Net) error rate of 4.8 remains above Qwen3.5-Omni-Plus's 3.7. Audio MultiChallenge improves from 57.6 to 71.5 over the previous generation, whereas VoiceBench and WildSpeech remain below both Qwen3.5-Omni-Plus and Gemini 3.8 Flash.

\subsubsection{Audio-Visual Capabilities}

\begin{table}[H]
\centering
\caption{\textbf{Audio-visual capability evaluation of Qwen3.8-Omni-Flash and baseline models.
The highest scores are shown in bold.}}
\vspace{-.1in}
\label{tab:audio_visual_text}
\begin{adjustbox}{width=.95\textwidth}
\begin{threeparttable}
\small
\begin{tabular}{@{}lccccc@{}}
\toprule
\multicolumn{1}{c}{%
\begin{tabular}[c]{@{}c@{}}
\textbf{Benchmark}
\end{tabular}}
& \begin{tabular}[c]{@{}c@{}}
\textbf{Qwen3.8-Omni-}\\
\textbf{Flash}
\end{tabular}
& \begin{tabular}[c]{@{}c@{}}
\textbf{Qwen3.5-Omni-}\\
\textbf{Plus}
\end{tabular}
& \begin{tabular}[c]{@{}c@{}}
\textbf{Gemini 3.8}\\
\textbf{Flash}
\end{tabular}
& \begin{tabular}[c]{@{}c@{}}
\textbf{Seed 2.0 Lite}
\end{tabular}
& \begin{tabular}[c]{@{}c@{}}
\textbf{Muse Spark 1.2}
\end{tabular}
\\ \midrule
\multicolumn{6}{c}{\textit{Audio-Visual Understanding}} \\ \midrule
DailyOmni        & \textbf{85.1} & \textbf{85.1} & 84.0          & 81.4          & 79.6          \\
WorldSense       & 68.5          & 63.9          & \textbf{69.6} & 67.3          & 65.0          \\
AVUT             & 86.6          & 85.9          & \textbf{88.0} & 81.5          & 82.4          \\
JoinAVBench      & \textbf{75.9} & 74.1          & 70.4          & 70.6          & 71.8          \\
AVSpeaker        & 77.2          & 71.1          & \textbf{83.9} & 75.8          & 69.9          \\
 \midrule
\multicolumn{6}{c}{\textit{Audio-Visual Reasoning}} \\ \midrule
OmniVideoBench   & 63.4          & 53.8          & \textbf{65.2} & 58.5          & 62.2          \\
Video-MME-v2     & 65.0          & 47.9          & \textbf{71.0} & 64.9          & --            \\
MMOU             & 78.8          & 68.7          & \textbf{84.5} & 67.0          & 78.8          \\ \midrule
\multicolumn{6}{c}{\textit{Long Audio-Visual Reasoning}} \\ \midrule
LVOmniBench      & 63.3          & 53.2          & \textbf{70.7} & --            & --            \\ \midrule
\multicolumn{6}{c}{\textit{Audio-Visual Caption}} \\ \midrule
OmniCloze        & 63.2          & 64.2          & 60.9          & 56.3          & \textbf{65.3} \\
OmniCap-IF
& 80.6 $\mid$ 28.2
& 72.1 $\mid$ 14.1
& \textbf{81.9} $\mid$ \textbf{28.3}
& 74.6 $\mid$ 18.1
& 77.9 $\mid$ 26.8 \\ \midrule
\multicolumn{6}{c}{\textit{Audio-Visual Interaction}} \\ \midrule
QIVD             & \textbf{69.6} & 65.6          & 69.1          & 62.0          & 62.0          \\
OmniVChat-Bench  & \textbf{89.0} & 52.1 & 65.6          & 45.2          & 69.8          \\
Omni2Web
& 56.7 $\mid$ \textbf{51.6}
& 29.6 $\mid$ 49.1
& \textbf{57.0} $\mid$ 50.8
& 28.3 $\mid$ 46.5
& 35.2 $\mid$ 27.6 \\ 
ODUbench
& \textbf{91.64} $\mid$ \textbf{89.57}
& 74.69 $\mid$ 69.62
& 72.06 $\mid$ 66.38
& 77.28 $\mid$ 67.87
& 76.87 $\mid$ 71.28 \\ 
ProactiveVideoQA & 61.5          & 35.1          & 58.0          & \textbf{68.3} & 63.6          \\
StreamingBench   & \textbf{80.8} & 57.1          & 79.9          & 77.2          & 77.8          \\
\midrule
\multicolumn{6}{c}{\textit{Audio-Visual Spatial Intelligence}} \\ \midrule
OmniEchoBench             & \textbf{43.9} & 23.3          & 32.0          & 28.3          & 31.3          \\ \bottomrule
\end{tabular}
\begin{tablenotes}[flushleft]
\footnotesize
\item[1] ODUbench results are reported as Audio $\mid$ VA;
the highest score for each setting is shown in bold.
\item[2] OmniCap-IF results are reported as CSR $\mid$ ISR;
the highest score for each metric is shown in bold.
\item[3] Omni2Web results are reported as Track A $\mid$ Track B;
the highest score for each track is shown in bold.
\end{tablenotes}
\end{threeparttable}
\end{adjustbox}
\vspace{-.1in}
\end{table}

We evaluate joint audio-visual understanding, reasoning, captioning, and interaction using the benchmarks in Table~\ref{tab:audio_visual_text}. The suite includes DailyOmni~\citep{dailyomni}, WorldSense~\citep{worldsense}, AVUT~\citep{avut}, AVSpeaker~\citep{avspeakerbench}, JoinAVBench~\citep{jointavbench}; OmniVideoBench~\citep{OmniVideoBench}, Video-MME-v2~\citep{Video-MME-v2}, MMOU~\citep{MMOU}, and LVOmniBench~\citep{lvomni-bench} for reasoning over video; OmniCloze~\citep{omnicloze} and OmniCap-IF~\citep{OmniCap-IF} for captioning; and QIVD~\citep{qivd}, ODUbench~\citep{chen2026omnidemandunderstandingbenchmark}, ProactiveVideoQA~\citep{ProactiveVideoQA}, StreamingBench~\citep{StreamingBench}, OmniVChat-Bench~\citep{he2026omnivchatsynthesizingbenchmarkingtraining}, and Omni2Web~\citep{han_2026_22855767} for interaction; OmniEchoBench~\citep{liuOmniEcho} for audiovisual spatial intelligence. The video-reasoning results provide the reference setting for the agentic comparison in Section~\ref{sec:agentic-omni-understanding}.

Compared with Qwen3.5-Omni-Plus, \method improves on OmniVideoBench from 53.8 to 63.4, Video-MME-v2 from 47.9 to 65.0, and LVOmniBench from 53.2 to 63.3. These results show stronger joint audio-visual performance across the evaluated tasks. Gemini 3.8 Flash nevertheless retains higher scores on the three video-reasoning benchmarks above, motivating the investigation of agentic inference rather than relying solely on direct input interpretation. Captioning and interaction also show substantial, though task-dependent, gains. On OmniCap-IF, CSR and ISR increase from 72.1 and 14.1 to 80.6 and 28.2, approaching Gemini 3.8 Flash's 81.9 and 28.3. StreamingBench improves from 57.1 to 80.8, and ProactiveVideoQA from 35.1 to 61.5. \method achieves the highest listed QIVD score of 69.6 and ties the previous generation on OmniVChat-Bench at 89.0. On Omni2Web, it scores 56.7 on Track A and 51.6 on Track B, compared with 29.6 and 49.1 previously. It also improves on both ODUbench settings significantly. In contrast, Seed 2.0 Lite remains ahead on ProactiveVideoQA. Beyond these tasks, we extend \method with omni-spatial intelligence, with its OmniEchoBench score increasing from 23.3 to 41.7.

\subsection{Omni Agentic Capabilities}
\label{sec:omni-agent-evaluation}
Building on these capability-specific evaluations, we focus on agent workflows that combine multimodal evidence with task-directed execution. We first evaluate multimodal tool use and web-search tasks, and then study whether an agent can improve audio-visual understanding by actively locating and verifying evidence. This section complements the text and visual agent tasks above by focusing on omni task execution and agentic perception.

\subsubsection{Multimodal Agent Benchmarks}
WildClawBench-MM~\citep{wildclawbench}, UniClawBench~\citep{uniclawbench}, and AgenticVBench~\citep{agenticvbench} assess multimodal tool use, while OmniGAIA~\citep{omnigaia} assesses tasks involving web search. WildClawBench-MM is restricted to WildClawBench tasks that involve images, video, or audio. WildClawBench-MM and AgenticVBench use Claude Code, UniClawBench uses OpenClaw, and OmniGAIA is evaluated without an external agent harness. These are distinct execution settings rather than a single shared harness.

\begin{table}[H]
\centering
\caption{\textbf{Omni agentic performance of Qwen3.8-Omni-Flash and baseline models.
The highest scores are shown in bold.}}
\vspace{-.1in}
\label{tab:omni_agentic}
\begin{adjustbox}{max width=\textwidth}
\begin{threeparttable}
\small
\begin{tabular}{@{}lcccc@{}}
\toprule
\multicolumn{1}{c}{%
\begin{tabular}[c]{@{}c@{}}
\textbf{Benchmark}
\end{tabular}}
& \begin{tabular}[c]{@{}c@{}}
\textbf{Qwen3.8-Omni-}\\
\textbf{Flash}
\end{tabular}
& \begin{tabular}[c]{@{}c@{}}
\textbf{Qwen3.5-Omni-}\\
\textbf{Plus}
\end{tabular}
& \begin{tabular}[c]{@{}c@{}}
\textbf{Gemini 3.8}\\
\textbf{Flash}
\end{tabular}
& \begin{tabular}[c]{@{}c@{}}
\textbf{Seed 2.0 Lite}
\end{tabular}
\\ \midrule
WildClawBench-MM & \textbf{71.0} & 34.5 & 58.9          & 41.9 \\
UniClawBench     & \textbf{69.6} & 67.1 & 69.0          & 61.2 \\
AgenticVBench    & 36.8          & 14.5 & \textbf{45.0} & 10.0 \\
OmniGAIA         & 74.0          & 57.2 & \textbf{78.6} & 64.4 \\ \bottomrule
\end{tabular}
\begin{tablenotes}[flushleft]
\footnotesize
\item[1] WildClawBench-MM: evaluated with Claude Code on the subset of
WildClawBench tasks involving images, video, or audio.
\item[2] UniClawBench: evaluated with OpenClaw.
\item[3] AgenticVBench: evaluated with Claude Code.
\item[4] OmniGAIA: evaluated without an external agent harness.
\end{tablenotes}
\end{threeparttable}
\end{adjustbox}
\end{table}

Table~\ref{tab:omni_agentic} shows improvements over Qwen3.5-Omni-Plus on all four benchmarks. WildClawBench-MM increases from 34.5 to 71.0, AgenticVBench from 14.5 to 36.8, and OmniGAIA from 57.2 to 74.0, corresponding to gains of 36.5, 22.3, and 16.8 points. UniClawBench improves from 67.1 to 69.6. \method leads the listed models on WildClawBench-MM and UniClawBench; Gemini 3.8 Flash leads on AgenticVBench and OmniGAIA with 45.0 and 78.6. The comparison demonstrates stronger multimodal agent execution than the previous generation while identifying remaining gaps relative to other models.

\subsubsection{Agentic Omni Understanding}
\label{sec:agentic-omni-understanding}
Long audio-visual inputs often distribute relevant evidence across multiple segments. We compare two settings on OmniVideoBench, Video-MME-v2, and LVOmniBench: \textbf{Static}, in which the model directly interprets the input, and \textbf{Qwen Code}, in which an agent plans its approach, invokes tools, and iteratively locates and verifies relevant evidence. We apply both settings to \method and Gemini 3.8 Flash to examine how agentic execution changes each model's performance.

\begin{table}[H]
\centering
\caption{\textbf{Agentic omni understanding performance under Static and Qwen Code settings.
The highest scores are shown in bold.}}
\vspace{-.1in}
\label{tab:agentic_omni_understanding}
\begin{adjustbox}{max width=\textwidth}
\begin{threeparttable}
\small
\begin{tabular}{@{}lcccc@{}}
\toprule
\multicolumn{1}{c}{%
\begin{tabular}[c]{@{}c@{}}
\textbf{Benchmark}
\end{tabular}}
& \begin{tabular}[c]{@{}c@{}}
\textbf{Qwen3.8-Omni-Flash}\\
\textbf{(Static)}
\end{tabular}
& \begin{tabular}[c]{@{}c@{}}
\textbf{Qwen3.8-Omni-Flash}\\
\textbf{(Qwen Code)}
\end{tabular}
& \begin{tabular}[c]{@{}c@{}}
\textbf{Gemini 3.8 Flash}\\
\textbf{(Static)}
\end{tabular}
& \begin{tabular}[c]{@{}c@{}}
\textbf{Gemini 3.8 Flash}\\
\textbf{(Qwen Code)}
\end{tabular}
\\ \midrule
OmniVideoBench & 63.4 & 67.8          & 65.2 & \textbf{70.1} \\
Video-MME-v2   & 65.0 & 71.3          & 71.0 & \textbf{72.7} \\
LVOmniBench    & 63.3 & \textbf{73.6} & 70.7 & 70.7          \\ \bottomrule
\end{tabular}
\begin{tablenotes}[flushleft]
\footnotesize
\item[1] Static denotes direct interpretation of the input.
Qwen Code denotes an agentic setting in which the model uses Qwen Code
to plan its approach, call tools, and progressively locate and verify evidence.
\end{tablenotes}
\end{threeparttable}
\end{adjustbox}
\end{table}

Table~\ref{tab:agentic_omni_understanding} shows that \method benefits from agentic execution on all three benchmarks: OmniVideoBench improves from 63.4 to 67.8, Video-MME-v2 from 65.0 to 71.3, and LVOmniBench from 63.3 to 73.6. The largest gain is on LVOmniBench, where the model moves from 7.4 points below Gemini 3.8 Flash in the Static setting to 2.9 points above it in the Qwen Code setting. Gemini 3.8 Flash also benefits on OmniVideoBench and Video-MME-v2, and retains the highest agentic scores on those two tasks. Thus, the benefit of agentic evidence gathering depends on both the model and the task.

\paragraph{Token efficiency.}
On OmniVideoBench, agentic understanding improves accuracy while reducing reported token consumption per query from 145,736 to 79,117, a reduction of approximately 45.7\% (Table~\ref{tab:agentic_omni_efficiency}). The agentic mode preserves context across turns. This result shows an improved accuracy--token trade-off in the evaluated setting; it does not establish a corresponding reduction in end-to-end latency or monetary cost.

\begin{table}[H]
\centering
\caption{\textbf{Accuracy and token consumption of \method on OmniVideoBench
under Static and agentic understanding settings.}}
\vspace{-.1in}
\label{tab:agentic_omni_efficiency}
\begin{adjustbox}{max width=\textwidth}
\begin{threeparttable}
\small
\begin{tabular}{@{}lcc@{}}
\toprule
\textbf{Metric} & \textbf{Static} & \textbf{Qwen Code} \\ \midrule
Accuracy ($\uparrow$) & 63.4 & \textbf{67.8} \\
Tokens per query ($\downarrow$) & 145,736 & \textbf{79,117} \\ \bottomrule
\end{tabular}
\begin{tablenotes}[flushleft]
\footnotesize
\item[1] Agentic understanding preserves context across turns.
\end{tablenotes}
\end{threeparttable}
\end{adjustbox}
\end{table}

\section{Realtime Interaction}

\begin{figure}[tbh]
    \centering
    \includegraphics[width=1.0\textwidth]{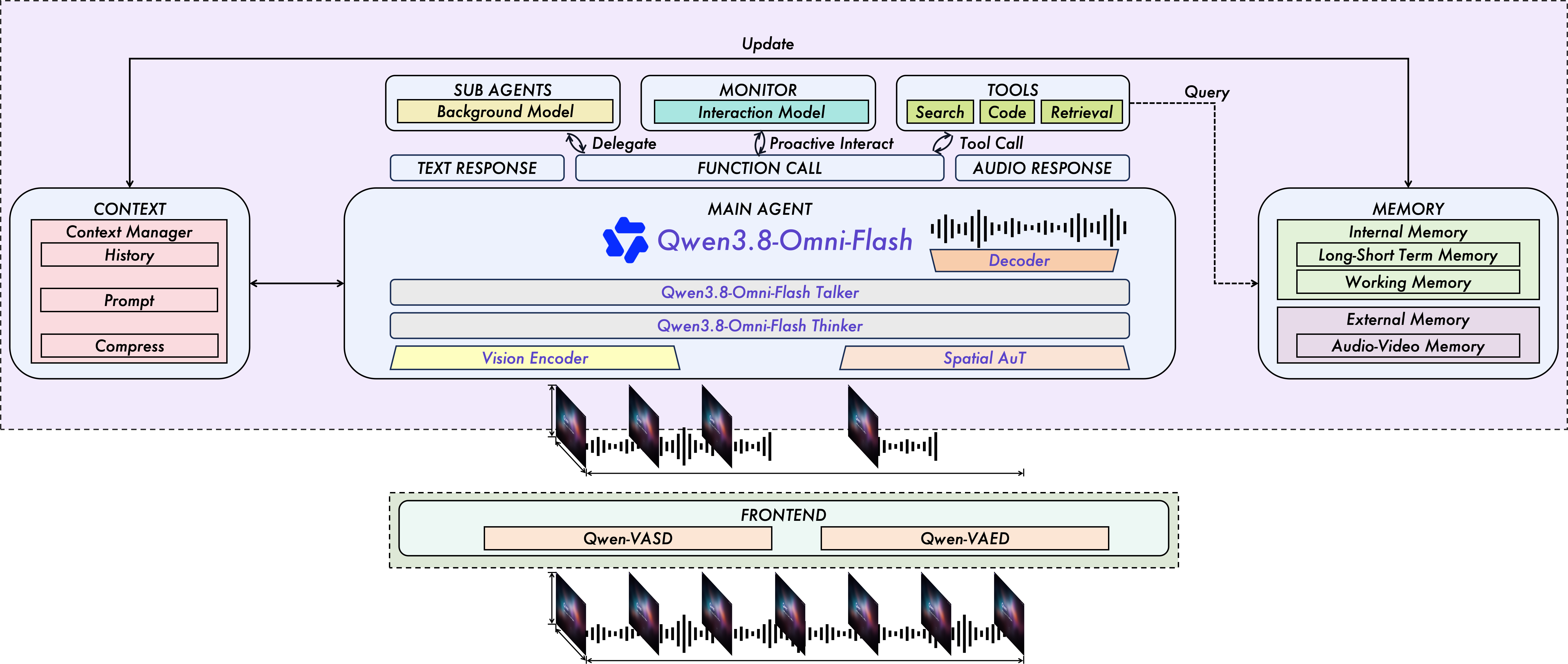}
    \caption{The overview of \realtime and Qwen-Live-Harness. \realtime adopts the Thinker-Talker architecture. Thinker is tasked with text generation while Talker focuses on generating streaming speech tokens by receives high-level representations directly from Thinker. To achieve ultra–low-latency streaming, Talker autoregressively predicts a multi-codebook sequence. At each decoding step, an MTP module outputs the residual codebooks for the current frame, after which the Code2Wav renderer incrementally synthesizes the corresponding waveform, enabling frame-by-frame streaming generation.}
    \label{fig:ovewview_realtime}
\end{figure}

Realtime omni-modal interaction requires a careful balance between responsiveness, speech naturalness, and agentic functionality. To address these challenges, we enhance the Talker module to generate more expressive and realistic spoken responses under stringent latency constraints, design an efficient inference pipeline to support low-latency realtime generation, and develop \textit{Qwen-Live-Harness} to extend \realtime to practical live interaction scenarios with asynchronous tool use, proactive interaction, and persistent memory. Together, as shown in Figure~\ref{fig:ovewview_realtime}, these components form a unified framework for realtime omni-modal applications.

\subsection{Talker}

Inherited from Qwen3.5-Omni-Plus~\citep{qwen35omni2026}, Talker operates on Residual Vector Quantization (RVQ) tokens and employs a Multi-Token Prediction (MTP) module to model residual codebooks, ensuring precise control over acoustic details. In addition, we design a dedicated system prompt to specify target voice characteristics and incorporate ARIA for streaming speech generation. Furthermore, an upsampling pathway is integrated into the causal Code2Wav decoder, elevating waveform reconstruction from 24 kHz to 48 kHz and substantially improving high-frequency fidelity.

We employ a five-stage training pipeline to enable Talker to adapt to the rich contextual representations from Thinker and generate natural spoken responses. In the initial pre-training stage, a dedicated data pipeline is established to build a comprehensive, balanced dataset. During continual pre-training (CPT), complex conversational context corpora are introduced to bolster Talker's contextual modeling capability. Following CPT, we train individual language experts and utilize Multi-Teacher On-Policy Distillation (MOPD)~\citep{MOPD} for multilingual speech generation, which effectively mitigates foreign accent artifacts induced by monolingual corpora. Subsequently, a lightweight speaker fine-tuning stage is conducted to capture target speaker characteristics. Finally, in the reinforcement learning stage, we fine-tune Thinker to align with human preferences and optimize Talker via GSPO~\citep{gspo} using reward signals from the aligned Thinker, further elevating overall performance.

\subsection{Efficiency}
\label{sec:realtime-efficiency}

To support low-latency inference, \realtime combines chunk-wise streaming input processing in Thinker with incremental speech generation in Talker. Talker consumes streamed Thinker outputs and uses ARIA to adaptively align and interleave text and speech tokens, enabling speech continuation from partial text prefixes. At each acoustic frame, a multi-token prediction (MTP) module predicts the residual RVQ codebooks, followed by a causal Code2Wav decoder that incrementally reconstructs the waveform. Together, these mechanisms allow text generation, speech-token prediction, and waveform reconstruction to proceed progressively, enabling audio delivery before the complete response has been generated.

\begin{table}[H]
    \centering
    \caption{\textbf{Realtime API efficiency of \realtime.}}
    \label{tab:realtime-api-efficiency}
    \small
    \setlength{\tabcolsep}{5pt}
    \begin{tabular}{@{}lccccc@{}}
        \toprule
        Input & Duration (s) & Text TPS & TTFT (ms) & TTFC (ms) & Generation RTF \\
        \midrule
        Audio       & 6  & 84.87 & 591.26 & 978.36  & 0.1538 \\
        Audio       & 12 & 83.18 & 604.80 & 982.74  & 0.1537 \\
        Audio       & 20 & 81.06 & 617.98 & 1026.39 & 0.1538 \\
        \midrule
        Audio-Video & 6  & 84.89 & 837.96 & 1214.73 & 0.1524 \\
        Audio-Video & 12 & 84.33 & 911.85 & 1268.07 & 0.1527 \\
        Audio-Video & 20 & 83.00 & 981.01 & 1350.49 & 0.1528 \\
        \bottomrule
    \end{tabular}
\end{table}

We evaluate the realtime API of \realtime using 6-, 12-, and 20-second audio-only and audio-video excerpts. Audio is transmitted as 16\,kHz mono PCM, with audio-video inputs additionally containing $1280\times720$ frames at 1 FPS. Each condition includes ten valid responses after one warm-up request, using fresh sessions and a fixed instruction requiring more than 200 output text tokens. Inputs are uploaded before explicit turn submission, with voice activity detection disabled.

Time-to-first-text (TTFT) and time-to-first-audio-chunk (TTFC) are measured from client-side input commitment to the first nonempty output text and audio chunks, respectively. They include network transit and server-side processing but exclude connection establishment, input upload, and playback. Text TPS measures streaming text throughput using server-reported token counts. Audio generation real-time factor (RTF) is the interval between the first and last audio chunks divided by the generated waveform duration. Both generation metrics exclude the initial response waiting time.

As shown in Table~\ref{tab:realtime-api-efficiency}, text throughput remains within 81.06--84.89 tokens per second. Reported first-audio latencies are approximately 0.98--1.03 seconds for audio-only inputs and 1.21--1.35 seconds for audio-video inputs. Visual input increases startup latency while sustained generation efficiency remains similar. Generation RTF stays near 0.153, corresponding to approximately $6.5\times$ realtime generation speed under the evaluated setting.

\subsection{Harness for Realtime Application}
\label{sec:realtime-harness}

To connect realtime multimodal interaction with practical agentic workflows, we develop \textit{Qwen-Live-Harness}. The framework connects \realtime to desktop audio-visual inputs and external agent backends, combining low-latency conversation with asynchronous execution, proactive interaction, and persistent memory.

\paragraph{Asynchronous tool use and sub-agent delegation.}
The foreground agent delegates tasks to external backends, including Qwen Code, Codex, and Claude Code, through a unified adapter interface. Task submission returns an immediate acknowledgment, allowing the conversation to continue while backend execution proceeds independently. Progress and results are incorporated into the conversational context asynchronously, with concise spoken notifications delivered around ongoing speech and playback. Users can interrupt the current spoken response without cancelling delegated work, and backend tasks can continue after the realtime call ends.

\paragraph{Proactive interaction.}
The harness supports user-defined conditions over audio, visual input, and time. Perceptual monitoring tasks run in independent realtime sessions and produce textual event reports, which the foreground agent converts into spoken notifications when the conversation permits. Cooldowns and suppression of consecutive positive detections limit redundant notifications. This design supports screen monitoring, environmental observation, and scheduled reminders during live sessions.

\paragraph{Persistent memory.}
The harness combines recorded dialogue, explicit working memory, and selected user facts retained across conversations. Reusable information is consolidated into persistent profiles and retrieved through lexical search with optional embedding-based matching. When enabled, visual memory stores textual observations derived from captured frames. These mechanisms preserve relevant preferences and context across realtime sessions, supporting continuity in both conversation and task delegation.

\subsection{Performance}

We evaluate \realtime against \plus, Gemini 3.8 Flash, Seed 2.0 Lite, and Muse Spark 1.2 on several representative omni-modal interaction benchmarks. As shown in Table~\ref{tab:realtime_benchmarks}, the model delivers strong interaction performance, despite operating in a non-thinking mode to prioritize responsiveness.


\begin{table}[H]
\centering
\caption{\textbf{Multimodal interaction performance comparison of \realtime and baseline models. The highest scores are shown in bold.}}
\vspace{-.1in}
\label{tab:realtime_benchmarks}
\begin{adjustbox}{max width=.85\textwidth}
\begin{threeparttable}
\small
\begin{tabular}{@{}lccccc@{}}
\toprule
\multicolumn{1}{c}{%
\begin{tabular}[c]{@{}c@{}}
\textbf{Benchmark}
\end{tabular}}
& \begin{tabular}[c]{@{}c@{}}
\textbf{Qwen3.8-Omni-}\\
\textbf{Flash-Realtime}
\end{tabular}
& \begin{tabular}[c]{@{}c@{}}
\textbf{Qwen3.5-Omni-}\\
\textbf{Plus}
\end{tabular}
& \begin{tabular}[c]{@{}c@{}}
\textbf{Gemini 3.8}\\
\textbf{Flash}
\end{tabular}
& \begin{tabular}[c]{@{}c@{}}
\textbf{Seed 2.0 Lite}
\end{tabular}
& \begin{tabular}[c]{@{}c@{}}
\textbf{Muse Spark 1.2}
\end{tabular}
\\ \midrule
\multicolumn{6}{c}{\textit{Audio-Visual Interaction}} \\ \midrule
QIVD & 66.8 & 65.6 & \textbf{69.1} & 62.0 & 62.0 \\
ProactiveVideoQA & 58.1 & 35.1 & 58.0 & \textbf{68.3} & 63.6 \\
StreamingBench & 78.9 & 57.1 & \textbf{79.9} & 77.2 & 77.8 \\
OmniVChat-Bench & \textbf{82.3} & 52.1 & 65.6 & 45.2 & 69.8 \\
Omni2Web
& 45.1 $\mid$ \textbf{57.0} & 29.6 $\mid$ 49.1 & \textbf{57.0} $\mid$ 50.8 & 28.3 $\mid$ 46.5 & 35.2 $\mid$ 27.6 \\
ODUbench & \textbf{90.2 $\mid$ 86.2} & 74.7 $\mid$ 69.6 & 72.1 $\mid$ 66.4 & 77.3 $\mid$ 67.9 & 76.9 $\mid$ 71.3 \\ \midrule
\multicolumn{6}{c}{\textit{Audio Interaction}} \\ \midrule
WildSpeech & 72.5 & 75.7 & \textbf{76.4} & 74.5 & 73.4 \\
VoiceBench & 88.8 & \textbf{92.9} & 92.3 & 84.1 & 79.8 \\
\bottomrule
\end{tabular}
\begin{tablenotes}[flushleft]
\footnotesize
\item[1] ODUbench results are reported as Audio $\mid$ VA;
the highest score for each setting is shown in bold.
\item[2] Omni2Web results are reported as Track A $\mid$ Track B; the highest score for each setting is shown in bold.
\end{tablenotes}
\end{threeparttable}
\end{adjustbox}
\vspace{-.1in}
\end{table}

In particular, it achieves substantial gains over Gemini 3.8 Flash on OmniVChat-Bench, ODUbench, and Omni2Web (Track B). On OmniVChat-Bench, which evaluates conversational ability in everyday interaction scenarios, the model obtains 82.3, significantly outperforming 65.6 from Gemini 3.8 Flash. This result highlights its strong omni-modal dialogue capability in practical user-facing settings. On ODUbench, which evaluates the understanding of user intent across omni-modal inputs, the model obtains the best listed scores in both the audio and audio-visual settings, demonstrating that it can accurately capture user intention even under realtime constraints. Meanwhile, Omni2Web (Track B) evaluates the understanding of GUI-based interactions, and the strong result indicates that \realtime can serve as a capable agent for GUI interaction, a key requirement of Qwen-Live-Harness. Overall, these results demonstrate that the model preserves strong interaction quality while maintaining realtime responsiveness, making it well suited for practical live agentic applications.

\renewcommand{\arraystretch}{1.2} 

\newcommand{\bench}[2]{\makecell[c]{\textbf{#1}\\[1pt]{\footnotesize\itshape #2}}}

\begin{table}[H]
\centering
\caption{\textbf{Voice cloning capability comparison. Benchmarks marked with $^{*}$ are in-house.}}
\label{tab:realtime_voice_clone}
\begin{adjustbox}{max width=\textwidth}
\begin{threeparttable}
\small
\begin{tabular}{@{}ccccc@{}}
\toprule
\begin{tabular}[c]{@{}c@{}}\textbf{Benchmark}\end{tabular}
& \begin{tabular}[c]{@{}c@{}}\textbf{Qwen3.8-Omni-}\\\textbf{Flash-Realtime}\end{tabular}
& \begin{tabular}[c]{@{}c@{}}\textbf{Qwen3.5-Omni-}\\\textbf{Plus}\end{tabular}
& \begin{tabular}[c]{@{}c@{}}\textbf{SeedAudio}\end{tabular}
& \begin{tabular}[c]{@{}c@{}}\textbf{MiniMax-Speech-2.8}\end{tabular}
\\ \midrule

\multicolumn{5}{c}{\textit{Content Stability}} \\ \midrule
\bench{SEED}{zh $\mid$ en $\mid$ hard} & \textbf{0.74} $\mid$ \textbf{0.89} $\mid$ \textbf{5.29} & 0.99 $\mid$ 1.26 $\mid$ 6.37 & 1.76 $\mid$ 1.02 $\mid$ 7.70 & 0.82 $\mid$ 0.99 $\mid$ 9.08 \\
\bench{Multilingual Voice Cloning}{30 lang} & \textbf{3.17} & 3.48 & 19.77 & 6.05 \\
\bench{Cross-Lingual Voice Cloning}{12 cl-lang} & \textbf{3.63} & 3.74 & 7.80 & 10.59 \\
\bench{Accented Voice Cloning$^{*}$}{Chinese dialects} & \textbf{5.03} & 5.39 & 7.58 & 7.48 \\
\bench{SpeechSuperClue$^{*}$}{ml $\mid$ cl} & \textbf{3.34} $\mid$ \textbf{4.00} & 6.23 $\mid$ 8.08 & 4.99 $\mid$ 11.55 & 3.95 $\mid$ 4.83 \\
\bench{SwanBench-Speech}{zh $\mid$ en} & 1.96 $\mid$ \textbf{2.37} & 1.76 $\mid$ 2.86 & 1.59 $\mid$ 3.46 & \textbf{1.57} $\mid$ 3.57 \\
\bench{LongSpeechGeneration$^{*}$}{zh $\mid$ en} & \textbf{1.74} $\mid$ \textbf{1.59} & 19.25 $\mid$ 4.52 & 7.71 $\mid$ 5.09 & 7.11 $\mid$ 6.51 \\ \midrule

\multicolumn{5}{c}{\textit{Timbre Consistency}} \\ \midrule
\bench{SEED}{zh $\mid$ en $\mid$ hard} & \textbf{0.805} $\mid$ \textbf{0.760} $\mid$ \textbf{0.787} & 0.741 $\mid$ 0.706 $\mid$ 0.716 & 0.796 $\mid$ 0.751 $\mid$ 0.776 & 0.777 $\mid$ 0.692 $\mid$ 0.754 \\
\bench{Multilingual Voice Cloning}{30 lang} & \textbf{0.845} & 0.789 & 0.840 & 0.792 \\
\bench{Cross-Lingual Voice Cloning}{12 cl-lang} & \textbf{0.709} & 0.524 & 0.699 & 0.657 \\
\bench{Accented Voice Cloning$^{*}$}{Chinese dialects} & \textbf{0.711} & 0.641 & 0.696 & 0.666 \\
\bench{SpeechSuperClue$^{*}$}{ml $\mid$ cl} & \textbf{0.763} $\mid$ \textbf{0.674} & 0.670 $\mid$ 0.570 & 0.747 $\mid$ 0.647 & 0.714 $\mid$ 0.571 \\
\bench{SwanBench-Speech}{zh $\mid$ en} & \textbf{0.828} $\mid$ \textbf{0.795} & 0.768 $\mid$ 0.728 & 0.799 $\mid$ 0.752 & 0.804 $\mid$ 0.763 \\
\bench{LongSpeechGeneration$^{*}$}{zh $\mid$ en} & \textbf{0.820} $\mid$ \textbf{0.768} & 0.597 $\mid$ 0.470 & 0.811 $\mid$ 0.739 & 0.811 $\mid$ 0.764 \\ \midrule

\multicolumn{5}{c}{\textit{Overall Cloning Score}} \\ \midrule
\bench{SpeechSuperClue$^{*}$}{ml $\mid$ cl} & \textbf{3.749} $\mid$ \textbf{3.306} & 3.253 $\mid$ 2.835 & 3.448 $\mid$ 2.756 & 3.589 $\mid$ 3.021 \\ \bottomrule
\end{tabular}
\begin{tablenotes}[flushleft]
\footnotesize
\item[1] Content stability is measured by WER/CER (lower is better).
\item[2] timbre consistency is
measured by speaker similarity (higher is better). 
\item[3] Multiple values are separated by $\mid$,
in the order of the subset shown under each benchmark.
\end{tablenotes}
\end{threeparttable}
\end{adjustbox}
\end{table}

To evaluate real-time speech response capabilities, we benchmark \realtime against leading industry baselines, including SeedAudio, MiniMax-Speech-2.8, Gemini 3.1 TTS Flash, SeedTTS 2.0, and ElevenLabs V3, across both public and in-house evaluation suites. As detailed in Table~\ref{tab:realtime_voice_clone} and Table~\ref{tab:realtime_custom_voice}, \realtime demonstrates superior content stability, timbre consistency, controllability, and naturalness.
\textbf{Voice Cloning}. In zero-shot voice cloning (Table~\ref{tab:realtime_voice_clone}), \realtime consistently achieves the best content stability (lowest WER/CER) and highest timbre similarity across the SEED, multilingual, cross-lingual, and accented speech benchmarks. Notably, for long-horizon generation, it substantially mitigates degradation, maintaining robust stability on both SwanBench-Speech and LongSpeechGeneration. In challenging in-the-wild evaluations (SpeechSuperClue), \realtime achieves overall cloning scores of 3.749 (multilingual) and 3.306 (cross-lingual), outperforming SeedAudio and MiniMax-Speech-2.8 by a clear margin.
\textbf{Custom Voice}. As shown in Table~\ref{tab:realtime_custom_voice}, \realtime maintains leading content stability, excelling in phonetic precision (93.3\% accuracy on PhonePronunciation) and ultra-long speech generation (1.83 WER and 0.693 similarity). For stylistic controllability, \realtime attains a 95.0\% success rate, closely matching Gemini 3.1 TTS Flash (97.5\%). Furthermore, in speech naturalness arena, \realtime achieves a competitive 72.7\% win rate, approaching Gemini 3.1 TTS Flash (74.2\%) while significantly outperforming MiniMax 2.8 HD (48.0\%), SeedTTS 2.0 (66.2\%), and ElevenLabs V3 (23.1\%).

\renewcommand{\arraystretch}{1.2} 


\begin{table}[H]
\centering
\caption{\textbf{Comprehensive Custom-Voice capability comparison. Benchmarks marked with $^{*}$ are in-house.}}
\label{tab:realtime_custom_voice}
\begin{adjustbox}{max width=\textwidth}
\begin{threeparttable}
\small
\begin{tabular}{@{}ccccccc@{}}
\toprule
\begin{tabular}[c]{@{}c@{}}\textbf{Benchmark}\end{tabular}
& \begin{tabular}[c]{@{}c@{}}\textbf{Qwen3.8-Omni-}\\\textbf{Flash-Realtime}\end{tabular}
& \begin{tabular}[c]{@{}c@{}}\textbf{Qwen3.5-Omni-}\\\textbf{Plus}\end{tabular}
& \begin{tabular}[c]{@{}c@{}}\textbf{Gemini 3.1}\\\textbf{TTS Flash}\end{tabular}
& \begin{tabular}[c]{@{}c@{}}\textbf{Minimax}\\\textbf{2.8 HD}\end{tabular}
& \begin{tabular}[c]{@{}c@{}}\textbf{SeedTTS}\\\textbf{2.0}\end{tabular}
& \begin{tabular}[c]{@{}c@{}}\textbf{ElevenLabs}\\\textbf{V3}\end{tabular}
\\ \midrule

\multicolumn{7}{c}{\textit{Content Stability}} \\ \midrule
\bench{SEED}{zh $\mid$ en $\mid$ hard} & \textbf{0.83} $\mid$ \textbf{0.93} $\mid$ \textbf{5.22} & 1.12 $\mid$ 2.54 $\mid$ 7.30 & 2.41 $\mid$ 1.48 $\mid$ 19.46 & 1.10 $\mid$ 1.69 $\mid$ 6.82 & 0.87 $\mid$ 1.41 $\mid$ 6.87 & 1.62 $\mid$ 0.95 $\mid$ 12.51 \\
\bench{Multilingual test set}{29 lang} & \textbf{3.16} & 5.94 & 5.90 & 16.31 & -- & 6.86 \\
\bench{SpeechSuperClue$^{*}$}{zh $\mid$ en} & \textbf{7.19} $\mid$ 7.55 & 10.38 $\mid$ 12.51 & 9.15 $\mid$ \textbf{6.39} & 7.91 $\mid$ 6.94 & 7.46 $\mid$ 7.90 & 8.39 $\mid$ 7.38 \\
\bench{PhonePronunciation$^{*}$}{Phone accuracy} & \textbf{93.3\%} & 61.3\% & 86.6\% & 62.7\% & 67.2\% & 78.0\% \\
\bench{LongSpeechGeneration$^{*}$}{wer $\mid$ sim} & \textbf{1.83} $\mid$ \textbf{0.693} & 6.30 $\mid$ 0.597 & -- & -- & -- & -- \\ \midrule

\multicolumn{7}{c}{\textit{Control Effectiveness}} \\ \midrule
\bench{SpeechSuperClue$^{*}$}{20-style effectiveness} & 95.0\% & 61.5\% & \textbf{97.5\%} & -- & -- & -- \\ \midrule

\multicolumn{7}{c}{\textit{Subjective Score / Naturalness}} \\ \midrule
\bench{SpeechSuperClue$^{*}$}{ABX win rate} & 72.7\% & 28.0\% & \textbf{74.2\%} & 48.0\% & 66.2\% & 23.1\% \\ \bottomrule
\end{tabular}
\begin{tablenotes}[flushleft]
\footnotesize
\item[1] Content Stability and Multilingual test set are reported as WER (lower is better).
\item[2] LongSpeechGeneration is reported as WER $\mid$ speaker similarity (lower / higher is better).
\item[3] For PhonePronunciation accuracy, control effectiveness, and the subjective ABX win rate, higher is better.
\item[4] Multiple values are separated by $\mid$, in the order of the subset shown under each benchmark.
\end{tablenotes}
\end{threeparttable}
\end{adjustbox}
\end{table}

\section{Conclusion}
\label{sec:conclusion}

We presented \method, a natively multimodal agentic model that advances omni models toward real-world multimodal productivity. Its native multimodal co-training strategy preserves strong text capabilities while facilitating the transfer of reasoning and agentic capabilities to audio and video tasks, complemented by a 1M context window that supports long-form understanding and long-horizon planning. Extensive evaluations demonstrate strong performance across multimodal understanding, reasoning, agentic execution, and video productivity, supporting its deployment as either a primary agent or a specialized sub-agent. To translate these capabilities into practical workflows, we introduced \textit{Qwen-MM-Plugins}, which integrates audiovisual capabilities into existing agent harnesses through information abstraction, selective content access, and reusable productivity tools. We further introduced \textit{Qwen-Live-Harness} to support responsive, real-time multimodal agents through coordinated context and memory management, tool use, and sub-agent delegation. Together, \method and these open-source frameworks highlight the value of jointly advancing model capabilities and system infrastructure, providing a practical foundation for multimodal agents that perform sustained reasoning, planning, and execution across research and production workflows.

\section{Ethics Statement}
\label{sec:ethics}

Users generating content with this model should ensure that all input data are lawfully obtained and used, and obtain authorization from the individuals whose voices are used. They must not infringe on individuals’ rights to their voices or other personality rights and must comply with applicable laws and regulations. Synthetic audio made available to others must be labeled as AI-generated as required by law. Model deployers and users must fulfill their respective legal responsibilities.

\clearpage
\bibliography{biblio}
\bibliographystyle{colm2024_conference}

\clearpage
\section{Authors}

\textbf{Core Contributors\footnotemark}
\begin{multicols}{6}
{\fontsize{9}{10}\selectfont
Bing Han\\
Baosong Yang\\
Dake Guo\\
Dayiheng Liu$^*$\\
Fei Huang\\
Hangrui Hu\\
Hongkun Hao\\
Hao Wang\\
Hui Wang\\
Junming Lin\\
Jin Xu$^*$\\
Keda Tao\\
Linhan Ma\\
Pei Zhang\\
Qize Yang\\
Ruiyang Xu\\
Shun Lei\\
Ting He\\
Xize Cheng\\
Xun Gong\\
Xian Shi\\
Xiong Wang\\
Xinfa Zhu\\
Xinyu Zhang\\
Xueyao Zhang\\
Yunfei Chu\\
Yuan Feng\\
Yangze Li\\
Yuanjun Lv\\
Yongqi Wang\\
Yue Wang\\
Yuxuan Wang\\
Yu Xi\\
Yifan Yang\\
Yang Zhang\\
Zhifang Guo\\
Zishan Guo\\
Ziyue Jiang\\
Zhanzhao Liu\\
Zhenxin Lei\\
Zhijun Wang\\
}
\end{multicols}
\footnotetext{Alphabetical order. * denotes the corresponding author.}

\textbf{Contributors\footnotemark[\value{footnote}]}
\begin{multicols}{6}
{\fontsize{9}{10}\selectfont
An Yang\\
Bohua Chen\\
Bingshen Mu\\
Bochao Mao\\
Buxiao Wu\\
Bin Zhang\\
Bo Zheng\\
Bohan Zhang\\
Chuqiao Kuang\\
Chengpeng Li\\
Chenhao Li\\
Chenyuhao Wen\\
Chenhan Yuan\\
Donghua Cai\\
Dehui Kong\\
Dunjie Lu\\
Feng Wang\\
Fan Zhong\\
Fan Zhou\\
Gang Cheng\\
Guangyu Yuan\\
Hongqing Chen\\
Haolin He\\
Hongcheng Liu\\
Haoyu Wang\\
Haiyang Xu\\
Hui Xu\\
Haiquan Zhao\\
Huaqing Zhang\\
Jie Huang\\
Jingbin Hu\\
Jiayi Leng\\
Jie Li\\
Jiongnan Liu\\
Jitong Liao\\
Ju Li\\
Jiahao Meng\\
Jiaxuan Peng\\
Jianhong Tu\\
Jiaming Zhou\\
Jianwei Zhang\\
Junhao Zheng\\
Keqin Chen\\
Kexin Huang\\
Kangdi Wang\\
Lianghao Deng\\
Lei Huang\\
Lei Xie\\
Lingchen Meng\\
Liangzuo Sun\\
Ling Wang\\
Laiwen Zheng\\
Leying Zhang\\
Meng Gao\\
Mianqiu Huang\\
Minghao Han\\
Mei Li\\
Mingfeng Xue\\
Man Yuan\\
Mingkun Yang\\
Muzhi Zhu\\
Na Ni\\
Pengfei Wang\\
Qibing Bai\\
Qi Chen\\
Qidong Huang\\
Que Shen\\
Ruixun Liu\\
Rui Men\\
Rong Zhang\\
Shuai Bai\\
Su Hao\\
Sibo Song\\
Songsong Shao\\
Tao Chen\\
Tianyi Tang\\
Wenxiang Guo\\
Wen Huang\\
Xie Chen\\
Xionghui Chen\\
Xudong Guo\\
Xiao Li\\
Xuejing Liu\\
Xiaokai Peng\\
Xingzhang Ren\\
Xuancheng Ren\\
Xipin Wei\\
Xuechun Wang\\
Xiaodong Xu\\
Xi Zhang\\
Yiheng Chen\\
Yizhong Cao\\
Yang Fan\\
Yuan Ge\\
Yufei He\\
Yang Liu\\
Yanpeng Li\\
Yi Lu\\
Yuxin Liu\\
Yuxuan Liu\\
Yongxing Ma\\
Yunfei Mao\\
Yang Su\\
Yuchong Sun\\
Yueran Song\\
Yixuan Wang\\
Yang Xu\\
Yifan Ye\\
Yinsong Yan\\
Yuhuan You\\
Yichang Zhang\\
Yinger Zhang\\
Zibo Bi\\
Zihan Liu\\
Zijian Lin\\
Zihan Qiu\\
Zhixiang Ruan\\
Zekun Wang\\
Zhiyong Wu\\
Zhenghao Xing\\
Zeyu Yang\\
Zhaoqing Zhu\\
Zhiyuan Zhu\\
}
\end{multicols}

\section*{Acknowledgements}
\label{sec:ack}

We thank the Qwen-Audio team for their discussions and technical contributions regarding voice timbre in Qwen3.8-Omni-Flash-Realtime.

\clearpage

\end{document}